\documentclass{article}

\usepackage{arxiv}
\usepackage{adjustbox}
\usepackage[utf8]{inputenc} % allow utf-8 input
\usepackage[T1]{fontenc}    % use 8-bit T1 fonts
\usepackage{hyperref}       % hyperlinks
\usepackage{url}            % simple URL typesetting
\usepackage{booktabs}       % professional-quality tables
\usepackage{amsfonts}       % blackboard math symbols
\usepackage{nicefrac}       % compact symbols for 1/2, etc.
\usepackage{microtype}      % microtypography
\usepackage{lipsum}		% Can be removed after putting your text content
\usepackage{graphicx}
\usepackage{natbib}
\usepackage{doi}
\usepackage{bm}
\usepackage{rotating}
\usepackage{tikz}
\usepackage{caption,color,float}

\usepackage[bb=px]{mathalfa} 
\usepackage[varvw]{newtxmath}       % selects Times Roman as basic font
\usepackage{amsmath}
\mathchardef\mhyphen="2D % Define a "math hyphen"
\usepackage{booktabs}
\usepackage{multirow}
\usepackage{comment}
\usepackage{diagbox}
\usepackage{soul}
\usepackage{fancyhdr}
\usepackage{wrapfig}
\usepackage{graphicx}

\usepackage{subcaption}

\title{Explainable, Interpretable \& Trustworthy AI for Intelligent Digital Twin: Case Study on Remaining Useful Life }

\author{ 	{Kazuma ~Kobayashi} \\
	Nuclear, Plasma \& Radiological Engineering\\
	University of Illinois at Urbana-Champaign\\
	Urnaba, IL 61801, USA \\
	\And
      {Syed Bahauddin ~Alam} \\
	Nuclear, Plasma \& Radiological Engineering\\
	University of Illinois at Urbana-Champaign\\
	Urnaba, IL 61801, USA \\
 }

\renewcommand{\shorttitle}{\textit{Explainable, Interpretable \& Trustworthy AI for  Digital Twin} }

\hypersetup{
pdftitle={A template for the arxiv style},
pdfsubject={q-bio.NC, q-bio.QM},
pdfauthor={David S.~Hippocampus, Elias D.~Striatum},
pdfkeywords={First keyword, Second keyword, More},
}

\begin{document}
\maketitle

\begin{abstract}
%Good
Artificial intelligence (AI) and Machine learning (ML) are increasingly used in energy and engineering systems, but these models must be fair, unbiased, and explainable. It is critical to have confidence in AI's trustworthiness. ML techniques have been useful in predicting important parameters and in improving model performance. However, for these AI techniques to be useful for making decisions, they need to be audited, accounted for, and easy to understand. Therefore, the use of explainable AI (XAI) and interpretable machine learning (IML) is crucial for the accurate prediction of prognostics, such as remaining useful life (RUL), in a  digital twin system, to make it intelligent while ensuring that the AI model is transparent in its decision-making processes and that the predictions it generates can be understood and trusted by users. By using AI that is explainable, interpretable, and trustworthy, intelligent digital twin systems can make more accurate predictions of RUL, leading to better maintenance and repair planning, and ultimately, improved system performance. The objective of this paper is to explain the ideas of XAI and IML and to justify the important role of AI/ML in the digital twin framework and components, which requires XAI to understand the prediction better. This paper explains the importance of XAI and IML in both local and global aspects to ensure the use of trustworthy AI/ML applications for RUL prediction. We used the RUL prediction for the XAI and IML studies and leveraged the integrated Python toolbox for interpretable machine learning~(PiML).

\end{abstract}

% keywords can be removed

\section{Introduction}
The Office of Nuclear Regulatory Research (United States Federal Agency) has initiated a research effort to determine the feasibility of using digital twins (DTs)  for nuclear power applications. It has been demonstrated that a DT, a~platform for risk-informed decision making, can be leveraged in a variety of engineering contexts~\cite{rahman2022leveraging}. AI/ML-driven DT has the potential to revolutionize the energy industry  ~\cite{elsheikh2019modeling,mudhsh2023modelling,ghandourah2023performance,elsheikh2023water} by facilitating risk-informed decision-making and streamlining high-performing simulations through the analysis of vast amounts of data. The ~National Research Council (NRC) has partnered with the Idaho National Laboratory (INL) and Oak Ridge National Laboratory (ORNL) to evaluate the regulatory viability of DTs as part of this project~\cite{yadav2021technical}, which is listed as a top priority in the "FY2021-23 NRC Planned Research Activities"~\cite{NRCfuture}. Additionally, in ~2020 and 2021, the~NRC sponsored seminars on DT to advance nuclear technologies, and ~in 2021, it asked 12 industries for their opinions on the possible use of DT-AI/ML in advanced nuclear energy~\cite{ma2022exploring}.

It is important to address that the DT-enabling technologies consist of four major branches \cite{yadav2021technical}, to which our previous works have contributed: (i) M\&S, covering uncertainty quantification~\cite{kumar2022multi,kumar2022recent,kobayashi2022practical}, sensitivity analysis~\cite{kobayashi2022uncertainty,kumar2021quantitative}, data analytics through AI/ML, physics-based models, and~data-informed surrogate modeling~\cite{kobayashi2022surrogate,kobayashi2022data}; (ii) advanced sensors and instrumentation;~(iii) signal processing~\cite{kabir2010non,kabir2010theory,kabir2010watermarking,kabir2010loss}; and~(iv) data and information management. Additionally, DT simulations face a set of unique challenges: (a) treatment of noisy or erroneous data; (b) overall data inconsistencies~\cite{yadav2021technical}; (c) uncertainty quantification and sensitivity. All these challenges lead to a barrier in ML explainability for the DT. However, explainability in ML and DT must be fully understood for risk analysis to ensure better decision making using DT.  In~this regard, effective and trustworthy ML algorithms are important for explainable AI (XAI) \cite{gade2019explainable}.

Generally speaking, XAI is important for several reasons~\cite{longo2020explainable,gade2019explainable}:  1. One of the main challenges with AI is building trust in its decisions. By~explaining the rationale behind a decision, XAI promotes trust by enabling people to comprehend the decision's justification and have faith in its accuracy~\cite{holzinger2018machine,gade2019explainable}. 2. XAI can help increase AI systems' transparency, which is important for ensuring they are used ethically and responsibly. 3.  International regulations~\cite{roos2020european} require organizations to explain automated decision-making processes. XAI can help organizations to meet these requirements. 4. XAI can improve the user's experience by explaining AI decisions. XAI can be used to debug and improve AI systems by providing insight into how they make decisions and identify areas for improvement. Overall, XAI is an important area of research and development in AI. It can help build trust, increase transparency, meet regulatory requirements, improve the user experience, and~aid in debugging and improving AI systems~\cite{xu2019explainable,hoffman2018metrics}.  Furthermore, recent research on XAI for DTs has focused on improving model interpretability and trustworthiness. However, integrating complex AI architectures with digital twins while retaining interpretability remains challenging. Though complex models may improve accuracy, current implementations struggle to balance accuracy and interoperability ~\cite{xu2019explainable,hoffman2018metrics}. Further research should develop techniques to simplify AI model interpretation without sacrificing performance. This could involve new algorithms that elucidate AI decision-making logic within digital twins. More empirical studies are also needed to demonstrate these emerging techniques across applications~\cite{holzinger2018machine,gade2019explainable}. Overall, while research is evolving towards accessible and trustworthy AI integration, balancing predictive power and transparency is an open challenge. Future work will likely produce sophisticated amalgamations of AI and digital twins, offering reliable and interpretable forecasts that build user trust.

Additionally, XAI plays an important role in DTs~\cite{angulo2019towards}, which are virtual representations of physical systems, as it can ensure that their decisions are transparent and understandable to others. DTs use data and AI algorithms to make decisions and perform tasks, and~XAI can provide insights into how these decisions are made and why certain actions are being taken. To ensure the ethical and responsible use of DT, building trust is crucial. Furthermore, XAI provides insight into DTs' functioning and identifies areas for improvement regarding debugging and improving them. As~a result, the~DT can become more accurate and reliable, and~its performance can be optimized. XAI is an important consideration for DTs, as it can improve accuracy and reliability, increase transparency, and~build~trust.

Furthermore, the use of explainable, interpretable, and trustworthy artificial intelligence (AI) is crucial for accurately predicting the remaining useful life in an intelligent DT system~\cite{cannarile2019evidential,he2021digital}. These characteristics ensure that the AI model is transparent in its decision-making processes and that the predictions it generates can be understood and trusted by users. Explainability allows users to understand the factors most important in the model's prediction, interpretability allows the model's predictions to be easily understood by non-technical users, and trustworthiness ensures that the model is not making predictions based on irrelevant or misleading features and is not biased. By~using AI that is explainable, interpretable, and~trustworthy, intelligent DT systems can make more accurate predictions of RUL, leading to better maintenance and repair planning, and ultimately, improved system performance~\cite{guo2021real,lim2020state}.

%%%%%%%%%%%%%%%%%%%%%%%%%%%%%%%%%%%%%%%%%%

%\section{Digital Twins, XAI, and Remaining Useful~Life}
%192-Leveraging Industry 4.0
Prognostics and health management (PHM) use DTs to monitor and manage system health utilizing statistical algorithms and ~models ~\cite{shao2021data,yang2021remaining}. PHM can conduct condition-based maintenance, make maintenance decisions, and anticipate future failures in advance, which improves reliability and lowers maintenance costs. PHM relies on remaining usable life (RUL) for predictive maintenance, production-plan modifications, component management, and ~other decisions~\cite{chen2020machine,shao2021data}. To ~prevent system failures, RUL estimates must be made before and after failures, which has increased the importance of remaining useful life forecast research~\cite{ma2020deep,shao2021data}. RUL prediction methods can be model-based or~data-based~\cite{yang2019remaining,shao2021data}. Model-based RUL prediction methods use a physical model to estimate the remaining useful life of an asset. A ~detailed model of the asset's deterioration is needed for accurate predictions~\cite{sutharssan2012prognostics}. Advances in state monitoring technology have made it possible to develop reliable prediction models for assets in complex operating environments. Data-based RUL estimation uses monitoring data to make predictions. This process involves collecting data, processing them, modeling degradation, and ~predicting the life of the asset~\cite{wen2021generalized,shao2021data}.

DT's technology can reflect, mimic, and~forecast the status of the operating physical system in real-time and is one of the best tools for RUL prediction~\cite{lim2020state,shao2021data,he2021digital,shao2021data}. It does so based on data collected from the physical system by DT~\cite{zhuang2021connotation,shao2021data}. By ~establishing real-time health evaluations at the component level, improved safety margins can be achieved with the help of accurate estimates of  RUL~\cite{DOD1}. The abilities of DTs to provide prescriptions at the component level~\cite{DOD1}: (a) enhances operational readiness by limiting needless maintenance activities and offering the opportunity to match removals with scheduled maintenance systematically; (b) extends the life of the majority of individual components, hence saving costs by eliminating unnecessary removals; (c) improves the efficiency of maintenance cycles by allowing for conditional upkeep of particular objects. Additionally, the~DT offers assistance for RUL prediction, including benefits such as extensive health information data and reliable health indicators~\cite{he2021digital}. RUL prediction needs to have both a DT-driven, physical-model-based and a virtual-model-based, accurate machine learning method~\cite{he2021digital}. As~a result, machine learning techniques for DT-driven RUL prediction are of utmost~significance.

XAI can be a useful tool to increase the reliability and accuracy of RUL predictions made by AI systems. The~RUL of physical assets, such as machinery or infrastructure, can be predicted more accurately using XAI. RUL prediction calculates how long a piece of equipment can operate before it needs to be upgraded or replaced. Planning maintenance and allocating resources depend on this. The factors influencing the RUL predictions made by AI algorithms can be understood using XAI. Imagine, for instance, that an AI system foresees that a specific failure mode will result in a shorter RUL for an asset. In that case, XAI can explain why and how this failure mode will impact the performance of the asset. This is beneficial for the following reasons: 1. XAI can increase confidence in the AI system and its predictions by explaining RUL predictions. This is crucial to ensure that the predictions are taken seriously and applied to guide decision-making. XAI can be used for debugging and accuracy improvement by pointing out biases or errors in an AI system's predictions. Users can learn how the AI system generates its predictions and how to use them by using XAI. Regulations may sometimes require businesses to give reasons for using automated decision-making tools, such as RUL predictions. Organizations can work with XAI to satisfy these needs, and ~DT, combined with XAI, can help achieve~that.

PHM of critical infrastructure relies heavily on the ability to predict RUL and make data-driven maintenance decisions accurately. The explainable and interpretable AI techniques presented in this paper demonstrate significant promise to address real-world challenges in PHM. By generating explainable predictions, these methods can build trust and confidence in the RUL forecasts, enabling adoption in risk-averse, high-reliability domains. The model-agnostic nature allows flexibility in implementation across diverse assets and systems. Explanations of RUL predictions can provide directly actionable maintenance insights by revealing the operating conditions and failure modes driving degradation. As infrastructure continues to age and monitorability improves with sensor proliferation, the ability to leverage explainable AI for asset health management and life extension will only grow in practical importance and impact. With further refinement and validation, the XAI methods explored here can play a pivotal role in transforming infrastructure maintenance from time-based to data-driven strategies.

Despite the promise of XAI techniques for improving digital twin prognostics, significant challenges remain. Obtaining reliable and complete operational data across diverse assets to train accurate models is an ongoing obstacle. Complex engineering systems exhibit multifaceted deterioration processes that are difficult to capture, even with sophisticated AI architectures. Translating model explanations into tangible, cost-effective maintenance insights requires further validation. Metrics for quantitatively evaluating explanation quality remain an open research question. User studies are needed to determine how explanations impact operator trust and actions. Safety considerations around reliance on AI systems with limited interpretability raise ethical concerns. Systematically propagating uncertainties from data through model development into explanations also represents an unsolved problem. XAI still lacks flexibility for users to probe why specific predictions were made.  
To ensure trustworthy AI/ML applications for DT and its components, the objective of this paper is to explain the ideas of XAI and interpretable ML (IML) and justify the important role of AI/ML in the digital twin framework and components, which requires XAI to understand the predictions better. This paper explains the importance of XAI and IML in both local and global aspects to ensure the use of trustworthy AI/ML applications for RUL prediction. We leveraged the integrated Python toolbox for interpretable machine learning (PiML) \cite{sudjianto2022piml}. Additionally, the~XAI and interpretability of this paper have followed the approaches by~\cite{sudjianto2022piml,sudjianto2021designing,sudjianto2020unwrapping,yang2021gami}.

%%%%%%%%%%%%%%%%%%%%%%

\section{Core Components of Digital Twins: ML Algorithms and the XAI Requirement for~DTs} \label{MLXAI}

This section explains recent developments by the authors in digital twin research with simple illustrations: (a) an update module in the digital twin for temporal synchronization on the fly (Section~\ref{11}), (b) a faster prediction module for operator learning (Section~\ref{22}), and~(c) a digital twin framework (Section~\ref{33}). This section then justifies the important role of AI/ML in the DT framework, which requires explainability to understand the prediction better.  The~next section discusses explainable and interpretable~AI.

\subsection{Update Module in a Digital~Twin} \label{11}
\begin{figure*}[!htbp]
    \centering
    \includegraphics[width=\textwidth]{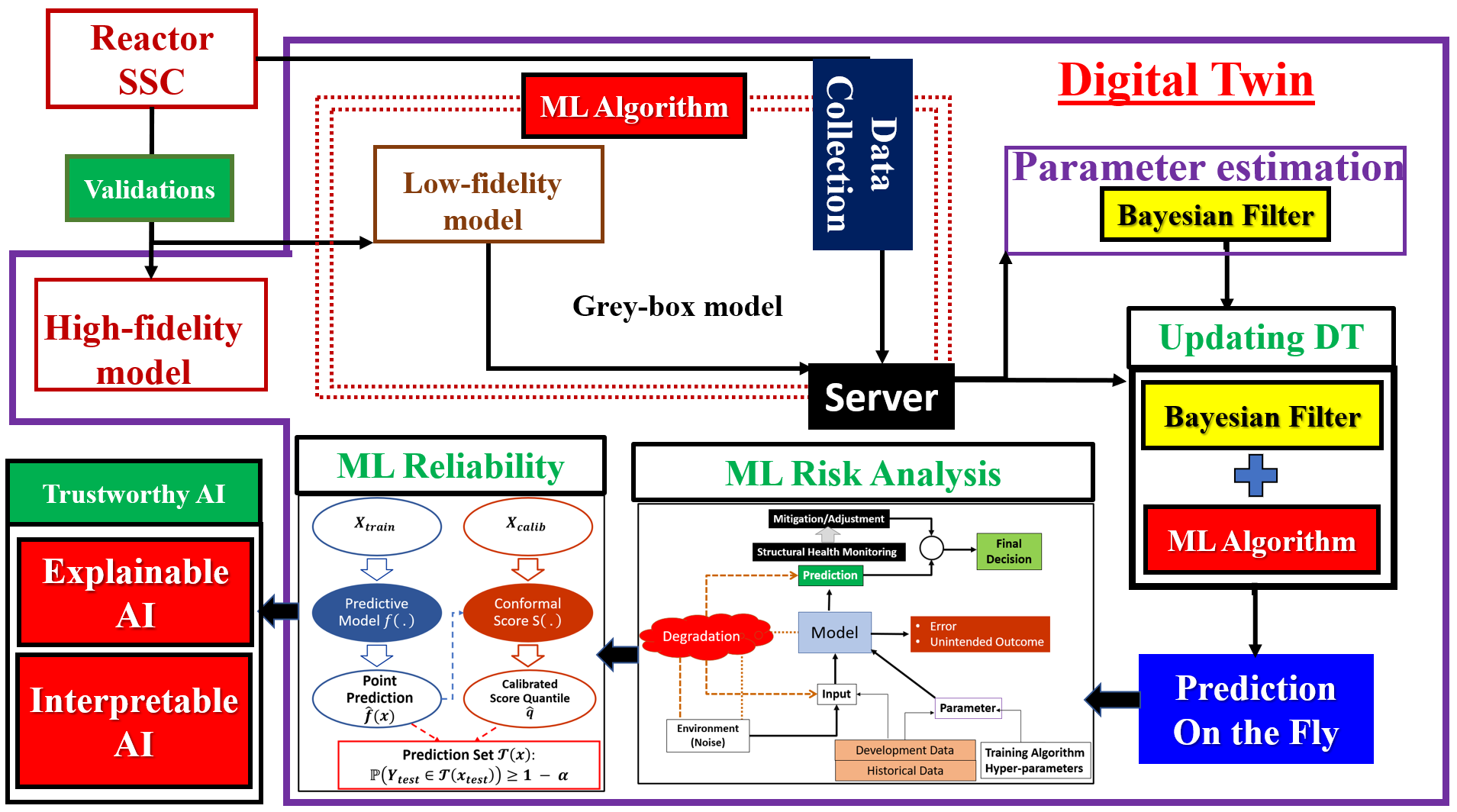}
    \caption{Intelligent digital twin framework with explainable AI and interpretable ML module. The diagram shows the ML components (red boxes) exploited in different segments of the digital twin framework for prognostics and justifies the explainability of these ML algorithms to understand RUL/prognostic behaviors ~\cite{kobayashi2023NET}.}
 \label{DTpic3}
\end{figure*}

There are five essential components needed to realize a DT framework: (1) the prediction module, (2) the system update module,  (3) the data processing module, (4) the visualization module, and~(5) the decision-making module. Among these modules, the prediction and system update ones require sophisticated ML algorithms. Additionally, the {system update module} is a key component that is not considered in ordinary simulations and is characteristic of digital twins. It leverages a Bayesian filter combined with an ML algorithm. Figure~\ref{DTpic3} shows the proposed intelligent DT framework with explainable AI and an interpretable ML module. This methodology has been adapted from the co-author Chakraborthy's previous work in ML-based DT~\cite{garg2021machine}. Figure~\ref{DTpic3} shows the ML components (red boxes) exploited in different segments of the DT framework for prognostics. Since different sophisticated ML algorithms are leveraged within an effective DT system, it is crucial to understand the explainability of these ML algorithms to better comprehend RUL/prognostic behaviors from the decision-making aspect. 

%\begin{comment}
%\end{comment}

The scope of this section is to explain the concept of a system in DT in terms of the application of ML---justify the XAI for DT update systems. The method for updating a DT  involves two approaches: (1) using the Bayesian filtering algorithm for estimating the parameters and states, and~(2) considering the temporal evolution of the system. To continuously update the model within the DT, online and sequential learning algorithms are necessary. 
 
The DT update module combines a Bayesian filter with a machine learning algorithm (Gaussian process) to predict the degradation mechanism as a future system state. It is important to have explainability and interpretability of the AI/ML systems used in practical DT systems, which will need to be updated over time~\cite{kobayashi2023NET}.

\begin{figure*}[!htbp]
    \centering
    \includegraphics[width=\textwidth]{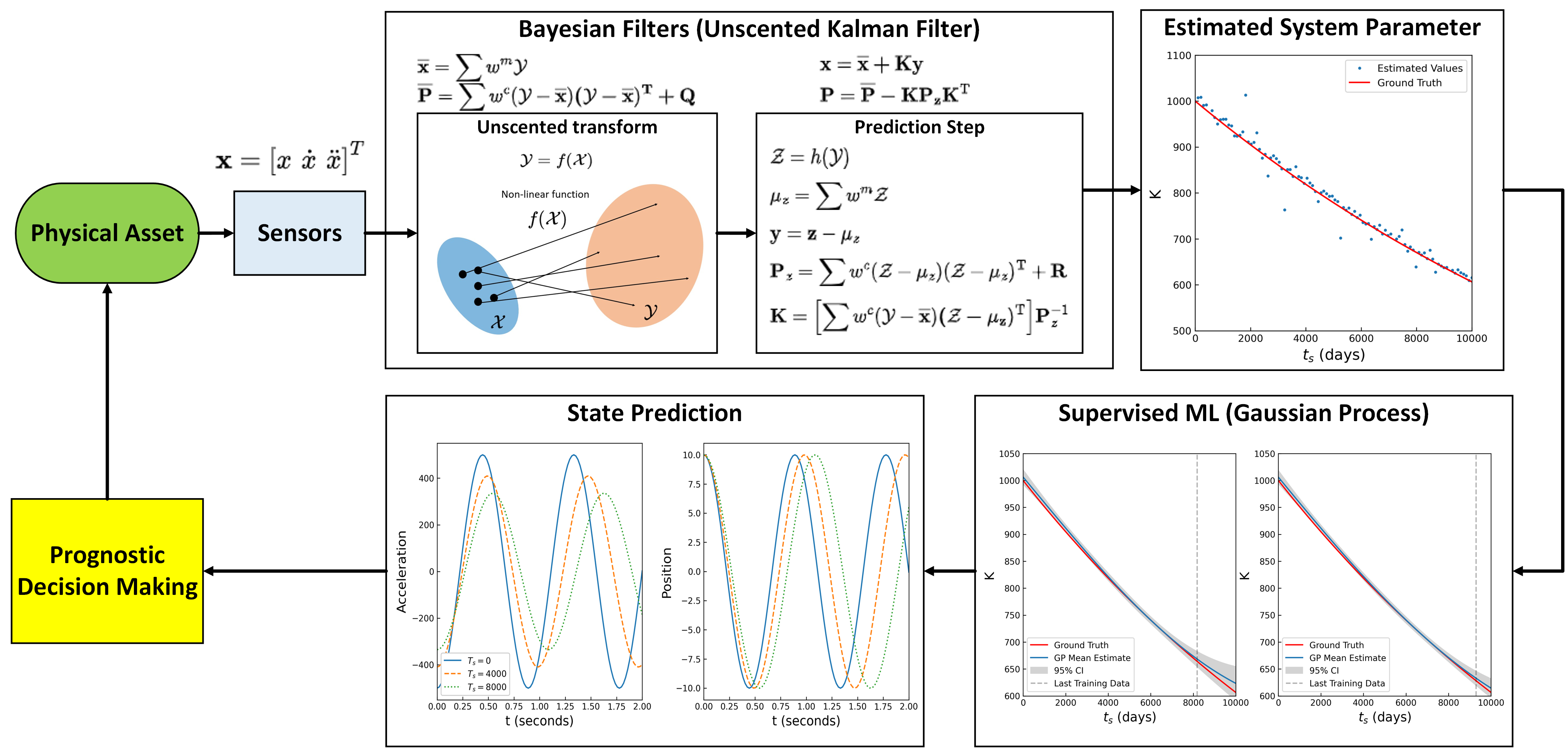}
    \caption{Developed update module  with an unscented Kalman filter (UKF) and ML method for an intelligent digital twin framework~\cite{kobayashi2023NET}.}
    \label{Update_Module}
\end{figure*}

Figure \ref{Update_Module} shows the methodology for updating system parameters with an unscented Kalman filter (UKF) combined with an ML model for Intelligent DT Framework, following the approach by co-author Chakraborty~\cite{garg2022physics,kobayashi2023NET}.

\subsection{Operator Learning as a Faster Surrogate for a Digital~Twin}
\label{22}

\begin{figure*}[!htbp]
    \centering
    \includegraphics[width=\textwidth]{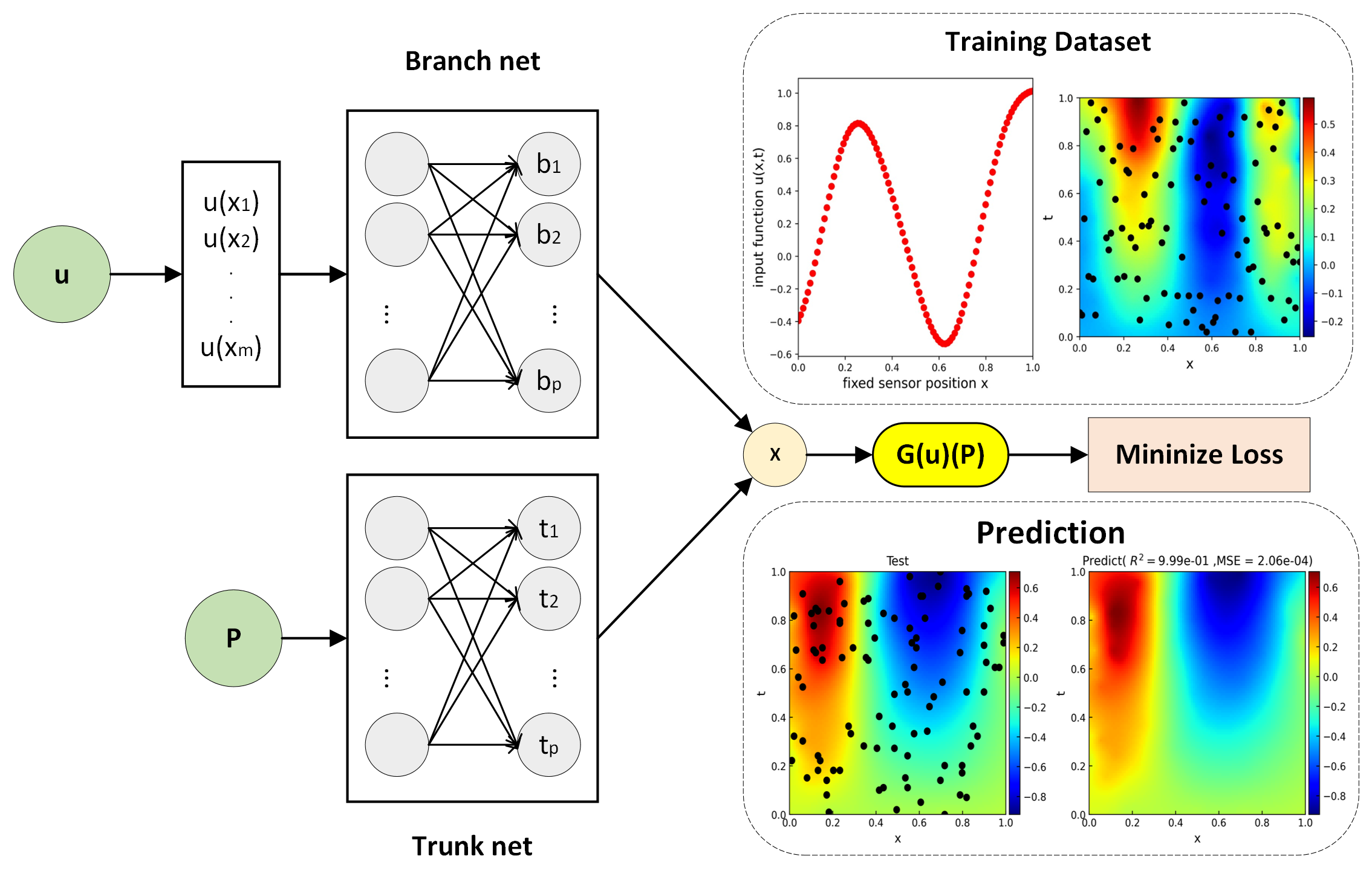}
    \caption{Authors leveraged the architecture of Deep Neural Operator (DeepONet) following the approach proposed by Lu~\cite{lu2021learning}. DeepONet comprises two networks: (1) Branch net and (2) Trunk net. The~case of a 2D diffusion system~\cite{kobayashi2023NET}.}
    \label{deeponet_process}
\end{figure*}

Figure~\ref{Update_Module} shows the methodology for updating the DT  system parameters for an intelligent DT framework. However, to ensure a practical and effective DT framework, one must achieve a good balance between prediction speed and accuracy. Although supervised ML algorithms can exhibit greater prediction accuracy, unfortunately, they are not considered a faster surrogate. Therefore, a state-of-the-art operator learning framework has proved to be a more robust and faster surrogate for the DT platform. Our preliminary studies developed an operator learning framework influenced by~\cite{lu2021learning,wang2021learning}, as~shown in Figure~\ref{deeponet_process}.
\begin{equation}
\label{eq:deeponet}
    \frac{ds(x)}{dx} = u(x), \,\,\, x \in (0, 1]
\end{equation}
where $s(x)$ is a target system state and $u(x)$ represents an input function. We will obtain the operator $G: u(x) \mapsto s(x)$ in this example. Following the approach~\cite{lu2022comprehensive}, the~input function $u(x)$ is sampled from a Gaussian random field at fixed $m$ positions: $x_{m}: \{x_{1},x_{2}, \cdots, x_{100} \}$. Since the solution of Equation~(\ref{eq:deeponet}) is required for training a model, $s(x)$ for each $u(x)$ at random location $P$ was numerically obtained. However, the main concern with operator learning~\cite{kobayashi2023NET} is its lack of~explainability.
  
\unskip  

\subsection{Platform-Agnostic and Surrogate-Driven Digital Twin~Framework }
\label{33}
\begin{figure*}[!htbp]
    \centering
    \includegraphics[width=\textwidth]{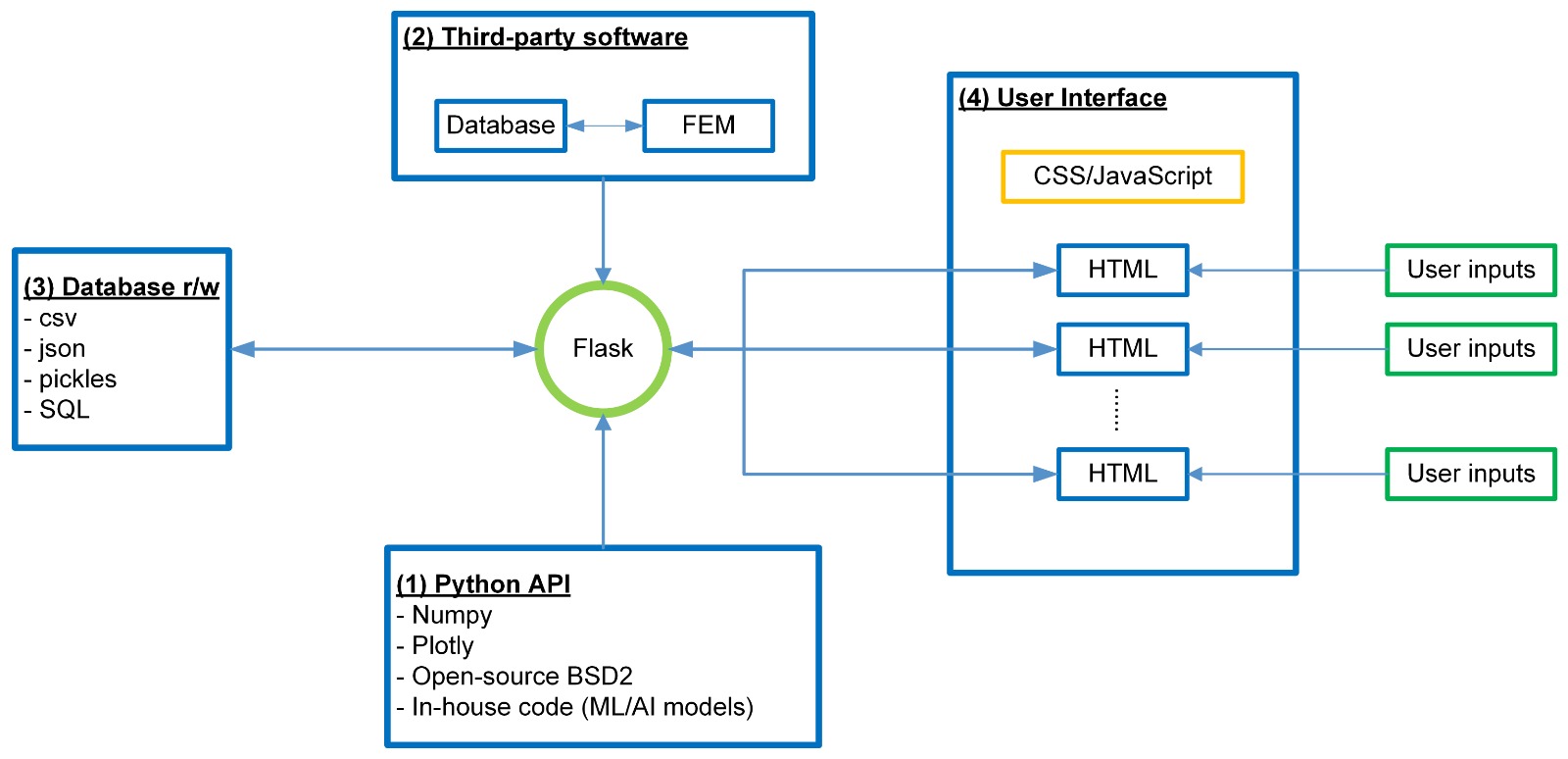}
    \caption{Developed architecture of DT framework following the approach by Bonney~\cite{bonney2021digital, bonney2022development}. The framework is composed of four parts: (1) Python application programming interface (API), (2) third-party software, (3) database, and~(4) user interface. Flask serves as the core for providing connectivity to all components.}
    \label{dt_platform_chart}
\end{figure*}

Commercial DT platforms are available from software vendors; however, they have several limitations: closed-source (modification and re-distribution are prohibited); framework (computer languages) and platform (operating system) dependent and expensive. Some of us have been developing in-house code following the approach by Bonney~\cite{bonney2021digital, bonney2022development}, as shown in Figure~\ref{dt_platform_chart}. This platform is platform-agnostic, generalizable, and modification-friendly; it has been used for applications that can modify/debug source code. This platform can be disseminated with collaborators and other researchers for simultaneous~improvement. 

This platform leverages Flask. It was used for developing the DT platform. Additionally, it is inside Python and uses all the associated libraries (numpy, scipy, Pandas, TensorFlow, PyTorch, matplotlib, plotly) inherent in Python. Additionally, Flask is a web application framework that can be handled with Python. It provides a graphical user interface (GUI) using a simple Python script and an HTML file~\cite{bonney2021digital, bonney2022development}. 

The connection between the database and the user is important in DT systems where the continuous passing of data from the physical system and simulation or ML model predictions based on those values is key. User inputs consisting of various data types (e.g., numerical, Boolean, textual) are parsed by the front-end interface into Python variables. These user-defined parameters serve to initialize and steer computational procedures and simulations within the platform. Concurrently, the platform leverages database functionalities, typically local databases (e.g., CSV files), to persistently store and retrieve simulation outcomes. This includes archiving results from user-driven simulations such as parametric sweeps in nonlinear continuation analyses. 

The synergistic interplay between user inputs and database operations facilitates a dynamic system where user-guided simulations can be reliably logged and accessed, establishing a robust foundation for subsequent computational workflows and inquiries. The integrated results, reflecting both user specifications and database contents, are then visualized to the user through the browser interface, typically using graphical and textual formats. This provides an informative output summarizing the computations. In summary, the bidirectional coupling between user inputs and database read/write operations supports the responsive and interactive nature of the DT platform, while enabling organized and performant data management.

\begin{figure*}[!htbp]
    \centering
    \includegraphics[width=\textwidth]{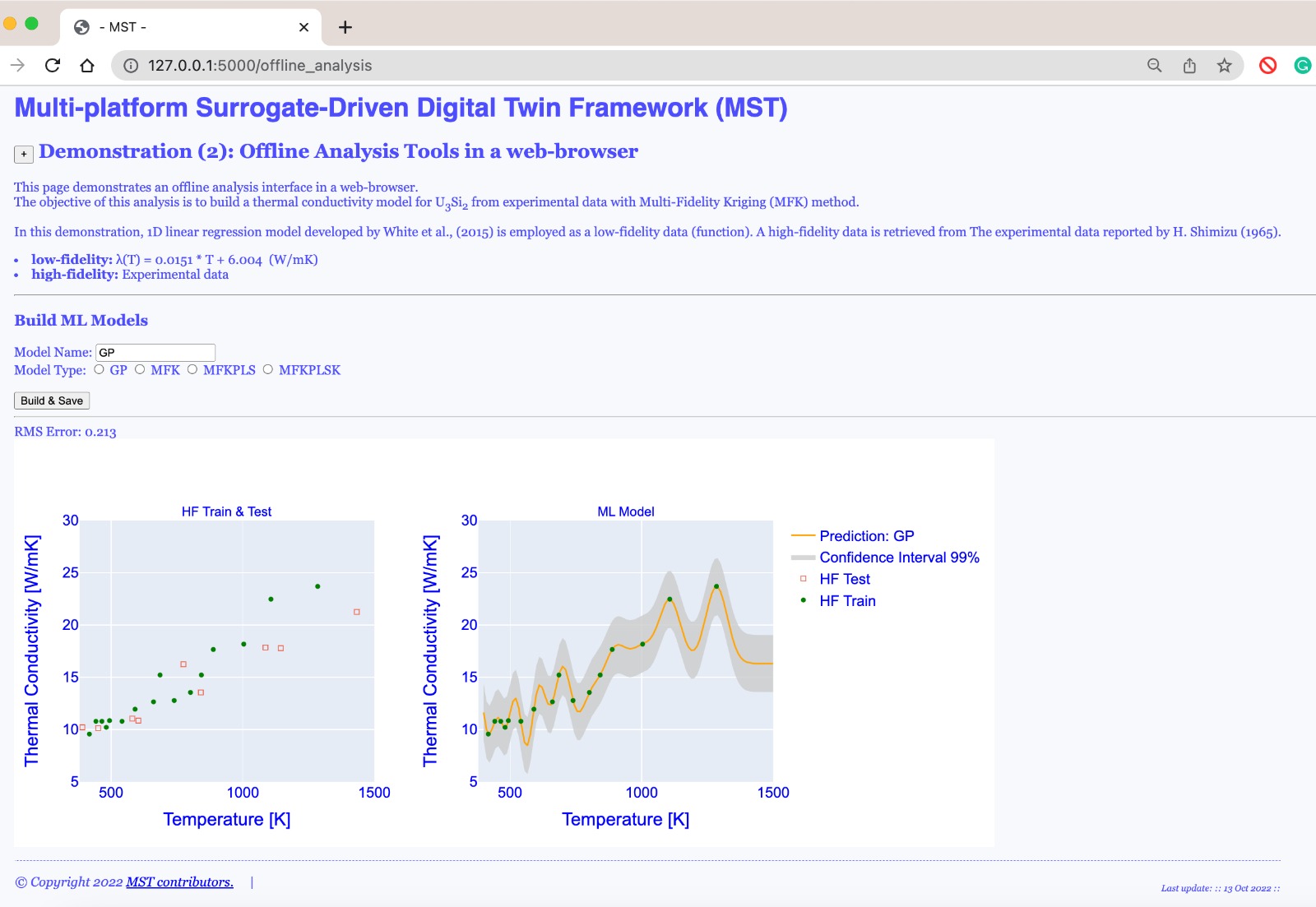}
    \caption{Demonstration of a Flask-based framework developed by the authors. Parameter setting, running Python code, and visualization can be performed on a web browser; following the approach by Bonney~\cite{bonney2021digital, bonney2022development}.}
    \label{dt_platform}
\end{figure*}

The architecture of DT framework proposed by Bonney~\cite{bonney2021digital, bonney2022development}. As seen in Figure~\ref{dt_platform}, the framework has four major components: (1) a Python application programming interface (API), (2) third-party software, (3) a database, and~(4) a user interface. Flask serves as the core for providing connectivity to all components.

\begin{itemize}
\item Python API: This component contains Python libraries such as visualization and mathematics. Additionally, our in-house AI/ML codes are classified.
\item  Third-party software: This category contains commercial or free third-party software.
\item Database: keep sensor data, simulation results, and models.
\item  User interface: graphical user interface through a web browser: click, drag and drop, etc.
\end{itemize}

\begin{comment}
    \begin{figure}[H]
\begin{adjustwidth}{-\extralength}{0cm}
\centering
\includegraphics[width=15.5cm]{dt_platform_chart.jpeg}
\end{adjustwidth}
\caption{Developed architecture of DT framework following the approach by Bonney~\cite{bonney2021digital, bonney2022development}. The framework is composed of four parts: (1) Python application programming interface (API), (2) third-party software, (3) database, and~(4) user interface. Flask serves as the core for providing connectivity to all components.}
    \label{dt_platform_chart}
\end{figure}
\end{comment}

\unskip  

\begin{comment}
    \begin{figure}[H]
\begin{adjustwidth}{-\extralength}{0cm}
\centering
\includegraphics[width=15.5cm]{dt_platform.jpeg}
\end{adjustwidth}
 \caption{Demonstration of a Flask-based framework developed by the authors. Parameter setting, running Python code, and visualization can be performed on a web browser.}
    \label{dt_platform}
\end{figure}
\end{comment}

%%%%
\section{Explainable and Interpretable AI for Digital~Twins}
\vspace{-3mm}
Section~\ref{MLXAI} discussed major components of the DT framework; however, they lack clear explainability--- in other words, XAI. Interpretable AI, which refers to the AI system's capacity to justify its judgments and predictions, is one of the key subfields of XAI. This is crucial because it enables users to comprehend the AI system's decision-making process and the rationale behind some of its actions. Developing AI systems that humans can understand requires a number of different strategies. One strategy is to employ transparent models, such as decision trees or linear models, which are simple to comprehend because they adhere to rules that are easily traced and explained. Another approach is to generate explanations after making a decision, which is known as post hoc explanations. This can be accomplished through the use of techniques such as feature importance or sensitivity analysis, which identify the factors that influenced the decision the most. AI systems need to be easy to understand so that people can trust them and use them ethically and responsibly. Insight into AI decision making and identifying potential improvement areas can also help debug and enhance AI systems. In conclusion, having interpretable AI can help with satisfying regulations and enhancing the user~experience.

Although XAI and interpretable AI are often used interchangeably to refer to the ability of the AI system to provide explanations for its decisions and predictions, there are some differences between XAI and interpretable AI. XAI means creating AI systems that can explain their actions, and~this can be accomplished through the use of transparent models and post hoc explanations, among other methods. In contrast, interpretable AI refers to interpreting and comprehending an AI model's outcomes. To do this, it may be necessary to extract meaningful patterns from the model, determine what factors lead to inaccurate predictions, or reveal sources of bias in the model. XAI is the broader concept of developing explainable and transparent AI systems, whereas interpretable AI is the process of deciphering and making sense of the outcomes of such systems.

\begin{figure}[H]
    \centering
    \includegraphics[scale=0.9]{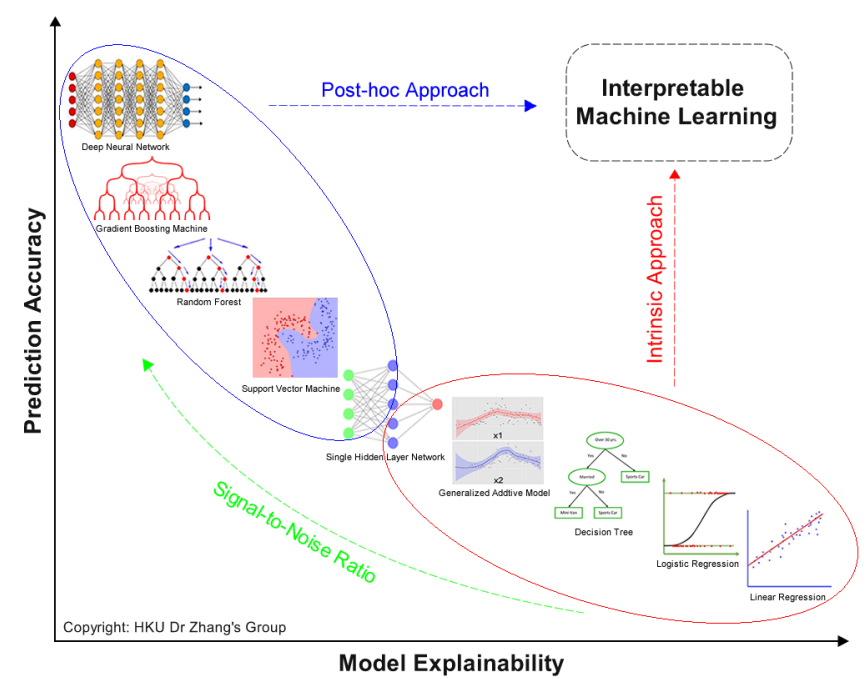}%MDPI: Please replace with sharper images
    \caption{Interpretable machine learning in terms of prediction accuracy vs. model explainability (sources: \cite{pimlModelInterpret1,STAT3612}).}
    \label{MLInterpret}
\end{figure}

Interpretable machine learning (IML) received substantial attention~\cite{9308260}. Model interpretation is the process of explaining how the input and output variables of a machine learning (ML) model are related to the decision-making process. It is important to increase users' trust in the system, but it may not always be possible to strengthen it due to legal limitations or the risk of unbiased conclusions. Human-readable justifications and explanations for the model results and predictions can be provided through interpretation procedures. Figure~\ref{MLInterpret} shows the diagram for interpretable machine learning in terms of prediction accuracy vs. model explainability~\cite{pimlModelInterpret1,STAT3612}.

There is a couple of popular interpretable methods~\cite{9308260} such as: 

\begin{itemize}
    \item ReLU-DNN~\cite{sudjianto2020unwrapping}.
    \item Explainable boosting machine (EBM) (\cite{nori2019interpretml,lou2013accurate}. 
    \item Fast interpretable greedy-tree sums (FIGS)  \cite{tan2022fast}.
    \item  Decision tree surrogates~\cite{Craven1995ExtractingTR}.
    \item  Partial dependence plot (PDP)  \cite{1013203451}.
    \item  Individual conditional expectation (ICE)  \cite{2014907095}.
    \item  Global surrogate~\cite{9308260}.
    \item  Explainable neural network (XNN) \cite{vaughan2018explainable}.
\end{itemize}

Instead of post hoc explainability models, this study uses inherently interpretable ML models: ReLU, EBM, FIGS, and~tree~\cite{pimlModelValidation2}.

%%%%%%%%%%%%%
\section{Inherent Interpretable ML Models: Problem Setup, Methodologies, and Data Preparation}

This study leverages the dataset from the 2008 Prognostics and Health Management Data Challenge (PHM08) \cite{sateesh2016deep,saxena2008damage,PHM08data,montoyalstm,montoyalstm}  to demonstrate the efficacy of XAI for RUL prediction. The PHM08 dataset was generated using the C-MAPSS simulation tool developed by NASA to mimic turbofan engine degradation under different operating conditions \cite{saxena2008damage}. Specifically, the simulation focused on modeling the degradation of the High-Pressure Compressor module across six combinations of altitude, throttle resolver angle, and Mach number \cite{saxena2008damage}. Damage propagation was allowed to continue until failure criteria were met in the simulation; the failure criteria for the PHM08 dataset were based on a health index that was defined as the minimum of several superimposed operational margins. These margins were determined by how far the engine was operating from various operational limits, such as stall and temperature limits. criterion was reached when the health index hit zero \cite{saxena2008damage}.

The PHM08 dataset contains sensor measurements representative of those typically used for real engine health monitoring. The data is structured as a matrix with discrete time intervals as rows and 26 feature columns \cite{PHM08data}. While the specific sensor types are not identified, the first few columns provide information on the engine identification number (ranging from 1 to 218), operational cycle, and 3 operating condition settings. The remaining 21 columns contain the core sensor data \cite{PHM08data}.

A key aspect of the original PHM08 data is the lack of explicit RUL values. To enable RUL prediction, we estimate RUL by subtracting the operational cycle from the maximum observed cycle for each engine. This approach aligns with common practices in prognostics where RUL is derived from operational data. Converting the raw data into a form suitable for RUL prediction is critical for applying and assessing XAI techniques in this domain. For demonstrative purposes in this study, only data from engine number 1 was utilized. As a result, 223 measurement times were available and divided at a ratio of 8:2 for training and testing of the ML models.

With the processed datasets, two XAI test cases were formulated for RUL prediction using the following interpretable machine learning models: (1) ReLU-DNN, (2) EBM, (3) FIGS, and (4) Decision Tree Surrogates. Details on the model implementation are provided in the following section.

%%%%%%%%%%%%%%%%%%%%%%%%%%%%%%%%%%%%%%%%%%
\section{XAI for RUL: Results and~Discussion}
This section aims to understand RUL prediction as our target explanation parameter. It is worth addressing that features consist of 26 independent values, s$_{i}$ represents sensor data, and ~setting$_{i}$  represents any operation setting. As addressed, we leveraged the integrated Python toolbox for interpretable machine learning (PiML) \cite{sudjianto2022piml}. Additionally, the~XAI and interpretability of this paper followed the approaches by~\cite{sudjianto2022piml,sudjianto2021designing,sudjianto2020unwrapping,yang2021gami}

\subsection{Feature~Selection}
\begin{figure}[H]
    \centering
    \includegraphics[width=\textwidth]{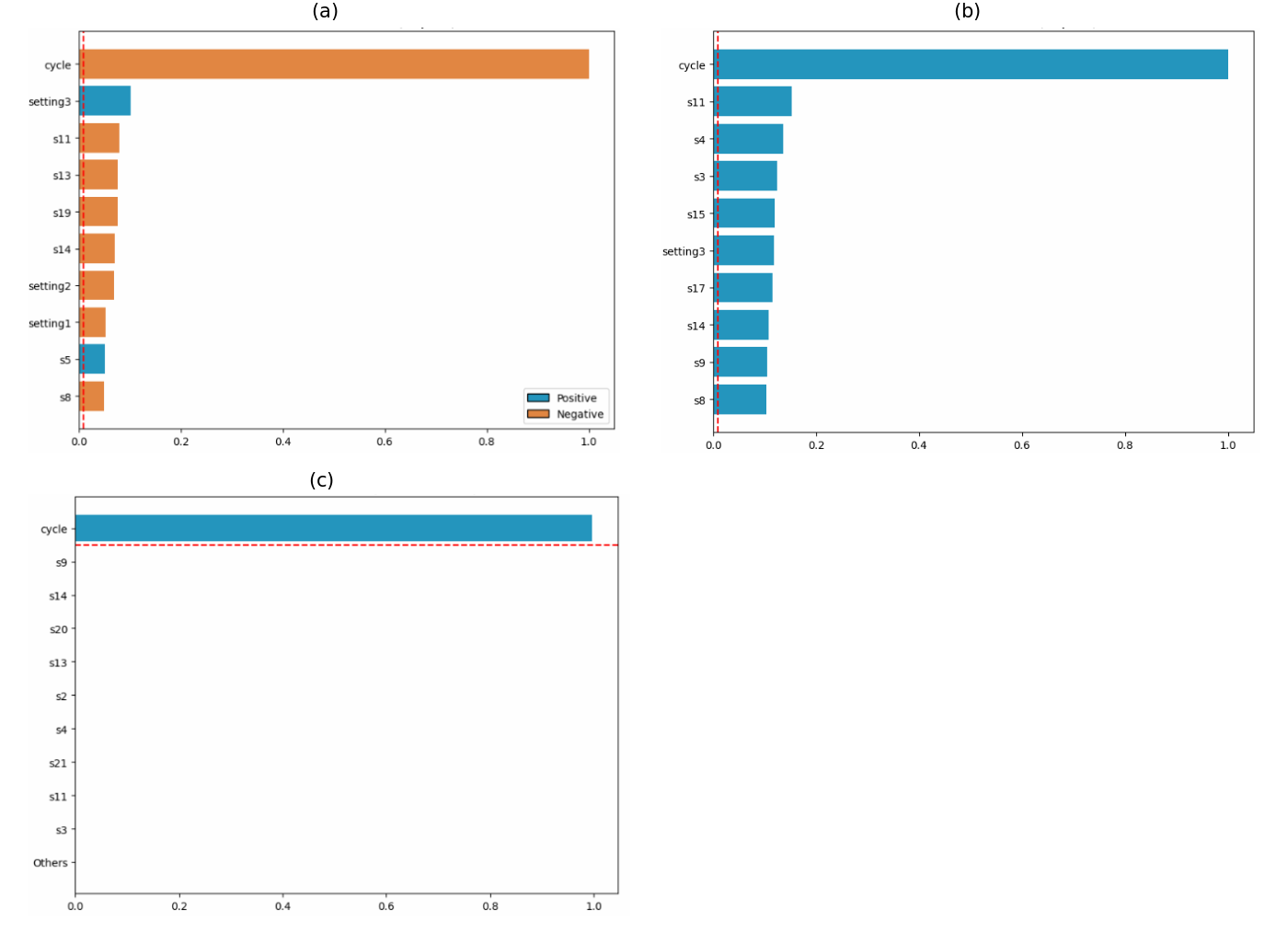}
    \caption{The features that exceed the threshold value of 0.01 was employed for building a model: (a) Pearson correlation coefficients of input features, (b) Distance correlation coefficients of input features, and (c) LGBM-based feature importance.}
    \label{fig:feature}
\end{figure}

%good
Feature selection reduces a large set of features to a manageable subset for machine learning model training. It is crucial to the ML process because it can improve the model's precision and interpretability. Feature selection can be used to explain XAI's post hoc decisions and predictions. Feature importance or sensitivity analysis can identify the model's most influential features. XAI feature selection is beneficial. It can reveal the model's decisions and the data's most important features. It can also identify and remove unnecessary features to improve the model's accuracy and reduce overfitting. Feature selection improves ML interpretability and transparency by showing how the model makes decisions. Figure~\ref{fig:feature} shows the feature selection in terms of  Pearson correlation (Figure~\ref{fig:feature}a), distance correlation (Figure~\ref{fig:feature}b), and feature importance (Figure~\ref{fig:feature}c). As shown in (Figure~\ref{fig:feature}a), "cycle operation" features most influence the model's decisions and mostly impact RUL~prediction.

\begin{comment}
\begin{figure}[H]
\begin{adjustwidth}{-\extralength}{0cm}
    \centering
    \includegraphics[scale=0.3]{Fig7.png}
    \caption{The features that exceed the threshold value of 0.01 was employed for building a model: (a) Pearson correlation coefficients of input features, (b) Distance correlation coefficients of input features, and (c) LGBM-based feature importance.}
    \label{fig:feature}
\end{adjustwidth}
\end{figure}    
\end{comment}

It is important to address that the Pearson's correlation  (Figure~\ref{fig:feature} a) is the most used metric for this purpose. Pearson correlation is a statistical technique for determining the linear relationship between two continuous variables. Pearson correlation can be used in feature selection to determine the features in a dataset that are most strongly linked with the target variable. The Pearson correlation coefficient is computed for each feature in the dataset with regard to the target variable to perform Pearson-correlation feature selection. The top-ranked features are then chosen to be included in the model after the features are rated according to their correlation coefficients. Pearson correlation for feature selection has the advantage of being a simple and fast method that is straightforward to implement. It is vital to keep in mind, though, that the Pearson correlation only identifies linear associations and might not be able to find non-linear interactions in the data.

The distance correlation (Figure~\ref{fig:feature}b) was used to quantify the relationship between two variables. The main objective of this correlation is to determine the most significant parameter for predicting the target variable while revealing non-linear relationships between attributes and the outcome. However, compared to the Pearson correlation, it is often more computationally demanding. Furthermore,  {light gradient boosting machine (LGBM)}-based feature importance was used for Figure~\ref{fig:feature}c. LGBM utilizes decision trees as its base model and trains multiple trees through an iterative process. As the trees are trained, the algorithm assigns higher weights to observations poorly predicted by previous trees, which helps improve the model's accuracy. LGBM has particular strengths in accuracy and the ability to handle large datasets while handling missing values and categorical features. It can be seen that both the distance correlation and LGBM-based feature importance confirm that "cycle operation" mostly impacts RUL prediction---which is consistent with the Pearson correlation.

\begin{figure}[H]
\centering
\subfloat[\centering]{\includegraphics[width=85mm]{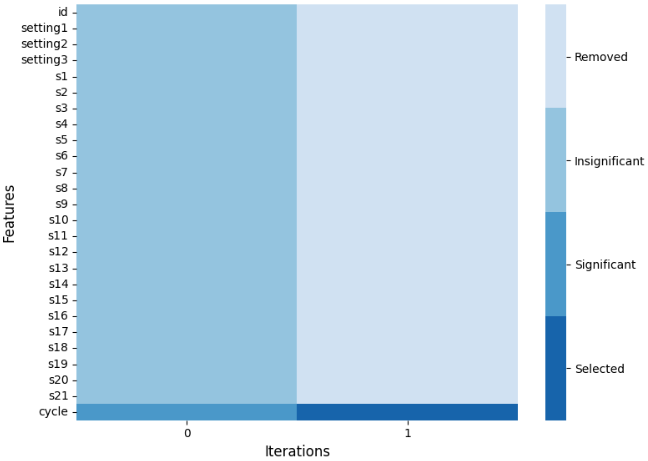}\label{cond}}
\subfloat[\centering]{\includegraphics[width=85mm]{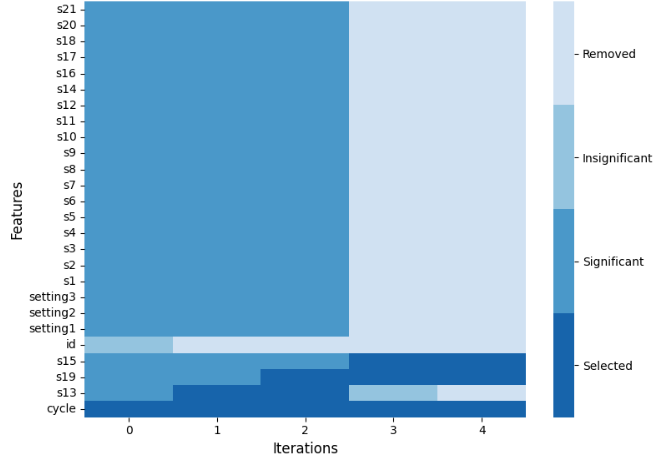}\label{condF}}
\caption{Conditional dependence using the relaxation randomized conditional independence test (RCIT) with a \emph{p}-value of 0.01 and a kernel size of 100 with different initializations:
(\textbf{a}) no initialization; 
(\textbf{b}) feature importance.}\label{fig:feature1}
\end{figure}

Furthermore, Figure~\ref{fig:feature} shows the conditional dependence using the relaxation randomized conditional independence test (RCIT). The RCIT  determines conditional independency for two variables given a set of other variables by drawing a random sample from the data. This can 
be valuable for estimating the relationships in the data and prediction. The procedure is repeated multiple times, and the results are averaged to obtain a final \emph{p}-value, indicating the likelihood that the two variables are dependent. However, RCIT is particularly effective for linear dependencies. 
Figure~\ref{fig:feature1}a,b shows that "cycle operation" mostly impacts RUL prediction, which is consistent with explanations from the feature selection in Figure~\ref{fig:feature}.  It is important to address that Pearson's correlation (Figure~\ref{fig:feature}a) is the most used metric for this purpose. For no initialization, Figure~\ref{fig:feature}a shows the conditional dependence between each feature and the target RUL value with no specific initialization method used. It highlights ``cycle" and other features as having higher dependence. On the other hand, for
feature importance initialization, Figure~\ref{fig:feature}b shows the conditional dependence after initializing the analysis using feature importance calculated separately. Initializing with feature importance focuses the conditional dependence estimation on the most relevant features for predicting RUL.

Pearson correlation is a statistical technique for determining the linear relationship between two continuous variables. Pearson correlation can be used in feature selection to determine the features in a dataset that are most strongly linked with the target variable. The Pearson correlation coefficient is computed for each feature in the dataset with regard to the target variable to perform Pearson-correlation feature selection. The top-ranked features are then chosen to be included in the model after the features are rated according to their correlation coefficients. Pearson correlation for feature selection has the advantage of being a simple and fast method that is straightforward to implement. It is vital to keep in mind, though, that the Pearson correlation only identifies linear associations and might not be able to find non-linear interactions in the data.

{The distance correlation} (Figure~\ref{fig:feature}b) was used to quantify the relationship between two variables. The main objective of this correlation is to determine the most significant parameter for predicting the target variable while revealing non-linear relationships between attributes and the outcome. However, compared to the Pearson correlation, it is often more computationally demanding. Furthermore,  {light gradient boosting machine (LGBM)}-based feature importance was used for Figure~\ref{fig:feature}c. LGBM utilizes decision trees as its base model and trains multiple trees through an iterative process. As the trees are trained, the algorithm assigns higher weights to observations poorly predicted by previous trees, which helps improve the model's accuracy. LGBM has particular strengths in accuracy and the ability to handle large datasets while handling missing values and categorical features. It can be seen that both the distance correlation and LGBM-based feature importance confirm that "cycle operation" mostly impacts RUL prediction---which is consistent with the Pearson correlation.

Furthermore, Figure~\ref{fig:feature} shows the conditional dependence using the relaxation randomized conditional independence test (RCIT). The RCIT  determines conditional independency for two variables given a set of other variables by drawing a random sample from the data. This can 
be valuable for estimating the relationships in the data and prediction. The procedure is repeated multiple times, and the results are averaged to obtain a final \emph{p}-value, indicating the likelihood that the two variables are dependent. However, RCIT is particularly effective for linear dependencies. 
Figure~\ref{fig:feature1}a,b shows that "cycle operation" mostly impacts RUL prediction, which is consistent with explanations from the feature selection in Figure~\ref{fig:feature}.

\begin{figure}[H]
\centering
%\begin{adjustwidth}{-1 in}{0cm}
\subfloat[\centering][\centering]{\includegraphics[width=70mm, height=70mm]{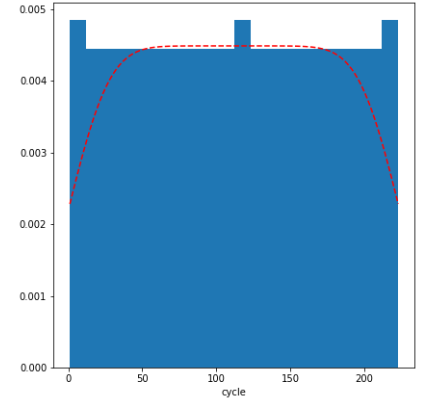}\label{univarCycle}}
\subfloat[\centering][\centering]{\includegraphics[width=70mm, height=70mm]{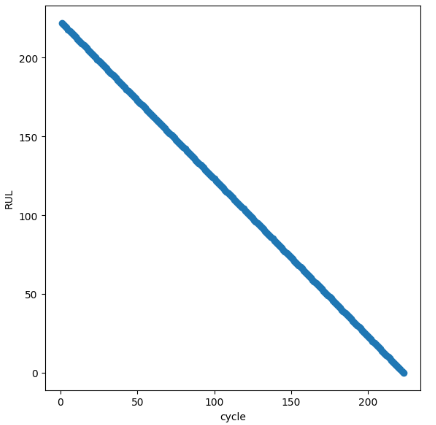}\label{CvR}}\\
\subfloat[\centering][\centering]{\includegraphics[width=70mm, height=70mm]{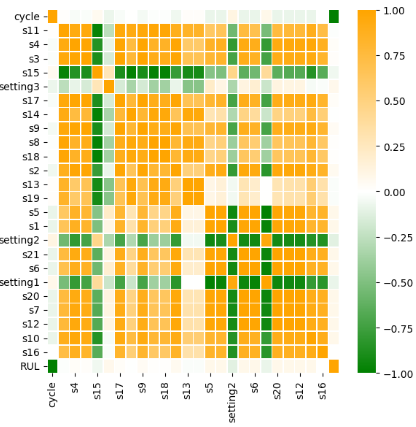}\label{heat}}
\caption{(\textbf{a}) Univariate  cycle operation; (\textbf{b}) Bivariate relationship between cycle operation and RUL. 
(\textbf{c}) Multivariate correlation heatmap.} \label{RULheat}
\end{figure}

Figure~\ref{RULheat} shows exploratory data analysis, check distribution, and correlation. Furthermore, it is confirmed by Figure~\ref{RULheat} that "cycle operation" impacts the RUL more than other operational input parameters. It is also important to explain the terms to the readers. Figure~\ref{RULheat}a shows the univariate cycle operation. It refers to the relationship between a single variable and a target or outcome variable. In other words, it is how the values of the single variable are related to the values of the target variable~\cite{demajo2020explainable}. Figure~\ref{RULheat}b shows the bivariate relationship between cycle operation and RUL. There are several types of bivariate relationships, including positive relationships, negative relationships, and no relationships. Figure~\ref{RULheat}c shows the multivariate correlation heatmap. It helps visualize correlations between variables and find data patterns. Each cell of a correlation heatmap typically represents the correlation between two variables, and cell color indicates correlation strength. Warmer colors, such as red and orange, indicate a stronger positive correlation; and cooler colors, such as blue, show a fairly strong negative correlation. The Pearson correlation coefficient for each pair of variables in a dataset is used to make a correlation heatmap~\cite{guyon2019cause}. The variables are plotted on the x and y axes in a matrix. The cells of the matrix are then color-coded according to the correlation's strength. 

\begin{comment}
Correlation heatmaps can quickly identify relationships between variables in a dataset and help understand its structure. Figure~\ref{RULheat}d shows the correlation graph with a cut-off at 20\%, which helps identify patterns in the data and understand how the variables are related.     
\end{comment}

\vspace{-12pt}

%%%%%%%%%%%%%%%%%%%%%%%%%%%%%%%%%%%%%%%%%%
XAI and interpretable ML model training refer to the process of building ML models that are easy to understand and explain. Several approaches can be used to train interpretable ML models, including simple, transparent models with small numbers of parameters and a simple structure---which are typically easier to interpret than more complex models. Examples include linear regression and decision trees. Regularization techniques can also be used to impose constraints on the model's complexity and prevent overfitting.  Feature selection can help identify and select the most important features in the data to improve the interpretability of the model by eliminating irrelevant or redundant features~\cite{longo2020explainable}. Examples include feature importance techniques and sensitivity analysis. Overall, interpretable ML model training is critical for improving the transparency and trustworthiness of ML models and providing insight into how the models' are making their decisions. Table~\ref{Modelconditions} shows the model customizations for the tree, FIGS, EBM, and~ReLU-DNN~models.

\begin{table}[htbp]
\centering
\caption{Trained model conditions.}
\label{Modelconditions}
\begin{adjustbox}{width=\textwidth}
\begin{tabular}{@{}llllllll@{}}
\toprule
\multicolumn{2}{c}{Tree}  & \multicolumn{2}{c}{FIGS} & \multicolumn{2}{c}{EBM}   & \multicolumn{2}{c}{ReLU-DNN}     \\ \midrule
Criterion & squared error & Max depth       & 5      & n interactions       & 10 & Layer size        & {[}40, 40{]} \\ \midrule
Max depth        & 5 & Max iterations & 100 & Outer bags        & 8      & Max epochs        & 1000  \\ \midrule
Min samples leaf & 5 &                &     & Inner bags        & 0      & Learning rate     & 0.001 \\ \midrule
Prune parameter  & 0 &                &     & Max bins          & 256    & Batch size        & 500   \\ \midrule
\multicolumn{4}{l}{\multirow{5}{*}{}}                & Max interaction bins & 32 & L1 regularization & 0.00001      \\ \cmidrule(l){5-8} 
\multicolumn{4}{l}{}                        & Max rounds        & 5000   & Dropout rate      & 0     \\ \cmidrule(l){5-8} 
\multicolumn{4}{l}{}                        & Early stop rounds & 50     & Early stop epochs & 20    \\ \cmidrule(l){5-8} 
\multicolumn{4}{l}{}                        & Early stop tol    & 0.0001 & Random state      & 0     \\ \cmidrule(l){5-8} 
\multicolumn{4}{l}{}                        & Random state      & 0      &                   &       \\ \bottomrule
\end{tabular}
\end{adjustbox}
\end{table}

Table~\ref{Modelperformance} shows the MAE, MSE, and~R2 performances of Tree, FIGS, EBM, and~ReLU-DNN models. It is clear that among all candidate models, ReLU-DNN exhibited superior performance, followed by EBM, FIGS, and~tree models, considering all the performance evaluations in the~table.

\begin{table}[!htbp]
\centering
\caption{Trained model performance.}
\label{Modelperformance}
\begin{adjustbox}{width=\textwidth}
\begin{tabular}{@{}llllllll@{}}
\toprule
Model    & Test MSE & Test MAE & Test R2 & Train MSE & Train MAE & Train R2 & Time \\ \midrule
ReLU-DNN & 0.0000   & 0.0025   & 0.9999  & 0.0000    & 0.0018    & 0.9999   & 4.6  \\
Tree     & 0.0001   & 0.0102   & 0.9980  & 0.0001    & 0.0083    & 0.9989   & 0.0  \\
FIGS     & 0.0002   & 0.0104   & 0.9977  & 0.0000    & 0.0015    & 0.9999   & 0.7  \\
EBM      & 0.0070   & 0.0607   & 0.9054  & 0.0003    & 0.0133    & 0.9961   & 3.8  \\ \bottomrule
\end{tabular}
\end{adjustbox}
\end{table}

\begin{comment}
\begin{table}[H]
  \caption{Trained model~performance.}
    \begin{adjustwidth}{-\extralength}{0cm}
  \newcolumntype{C}{>{\centering\arraybackslash}X}
\newcolumntype{L}{>{\raggedright\arraybackslash}X}
\newcolumntype{R}{>{\raggedleft\arraybackslash}X}
    \begin{tabularx}{\fulllength}{LLLLLLLL}
    \toprule
    \textbf{Model} & \multicolumn{1}{l}{\textbf{Test MSE}} & \multicolumn{1}{l}{\textbf{Test MAE}} & \multicolumn{1}{l}{\textbf{Test R2}} & \multicolumn{1}{l}{\textbf{Train MSE}} & \multicolumn{1}{l}{\textbf{Train MAE}} & \multicolumn{1}{l}{\textbf{Train R2}} & \multicolumn{1}{l}{\textbf{Time}}\\
    \midrule
    ReLU-DNN & 0     & 0.0025 & 0.9999 & 0     & 0.0018 & 0.9999 & 4.6\\
    \midrule
    Tree  & 0.0001 & 0.0102 & 0.998 & 0.0001 & 0.0083 & 0.9989 & 0\\
    \midrule
    FIGS  & 0.0002 & 0.0104 & 0.9977 & 0     & 0.0015 & 0.9999 & 0.7\\
    \midrule
    EBM   & 0.007 & 0.0607 & 0.9054 & 0.0003 & 0.0133 & 0.9961 & 3.8\\
    \bottomrule
    \end{tabularx}%
  \label{Modelperformance}%
    \end{adjustwidth}
\end{table}%

\end{comment}

\begin{figure}[H]
\centering
% \begin{adjustwidth}{-1in}{-1in} % Adjusts the left and right margins equally
% \nointerlineskip\leavevmode
\centering % Centers the content inside the adjustwidth environment
\subfloat[\centering]{\includegraphics[width=85mm]{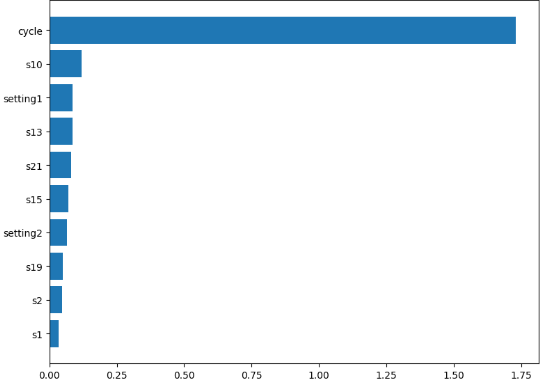}\label{reluGLOBAL}}
\subfloat[\centering]{\includegraphics[width=85mm]{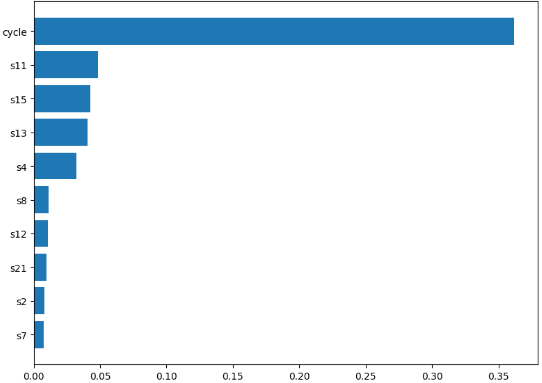}\label{ebmGLOBAL}}\\
\subfloat[\centering]{\includegraphics[width=85mm]{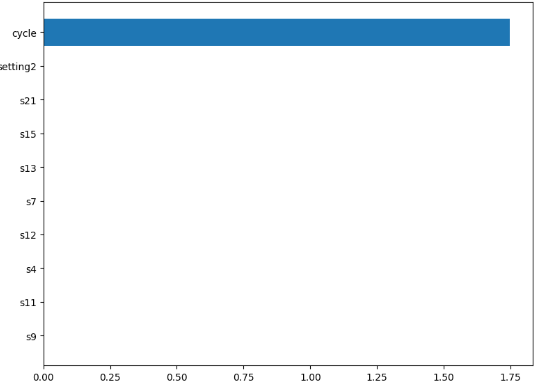}\label{figsGLOBAL}}
\subfloat[\centering]{\includegraphics[width=85mm]{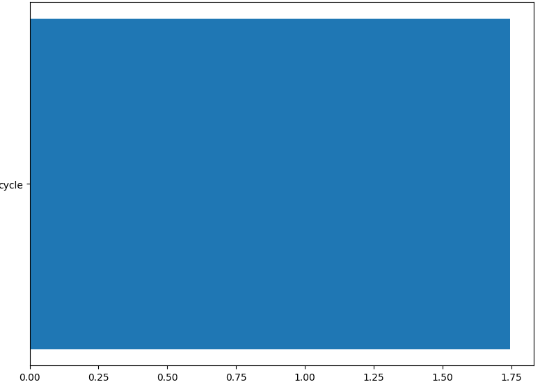}\label{treeGLOBAL}}
%\end{adjustwidth}
\caption{Permutation feature importance (PFI) for global model explainability:
(\textbf{a}) ReLU-DNN; 
(\textbf{b}) EBM;
(\textbf{c}) FIGS;
(\textbf{d}) Tree.}\label{ModelExplainGlobal}
\end{figure}

\subsubsection{Global Explainability:  Partial Dependence~Plot}
\begin{figure}[H]
\centering
\includegraphics[scale=0.7]{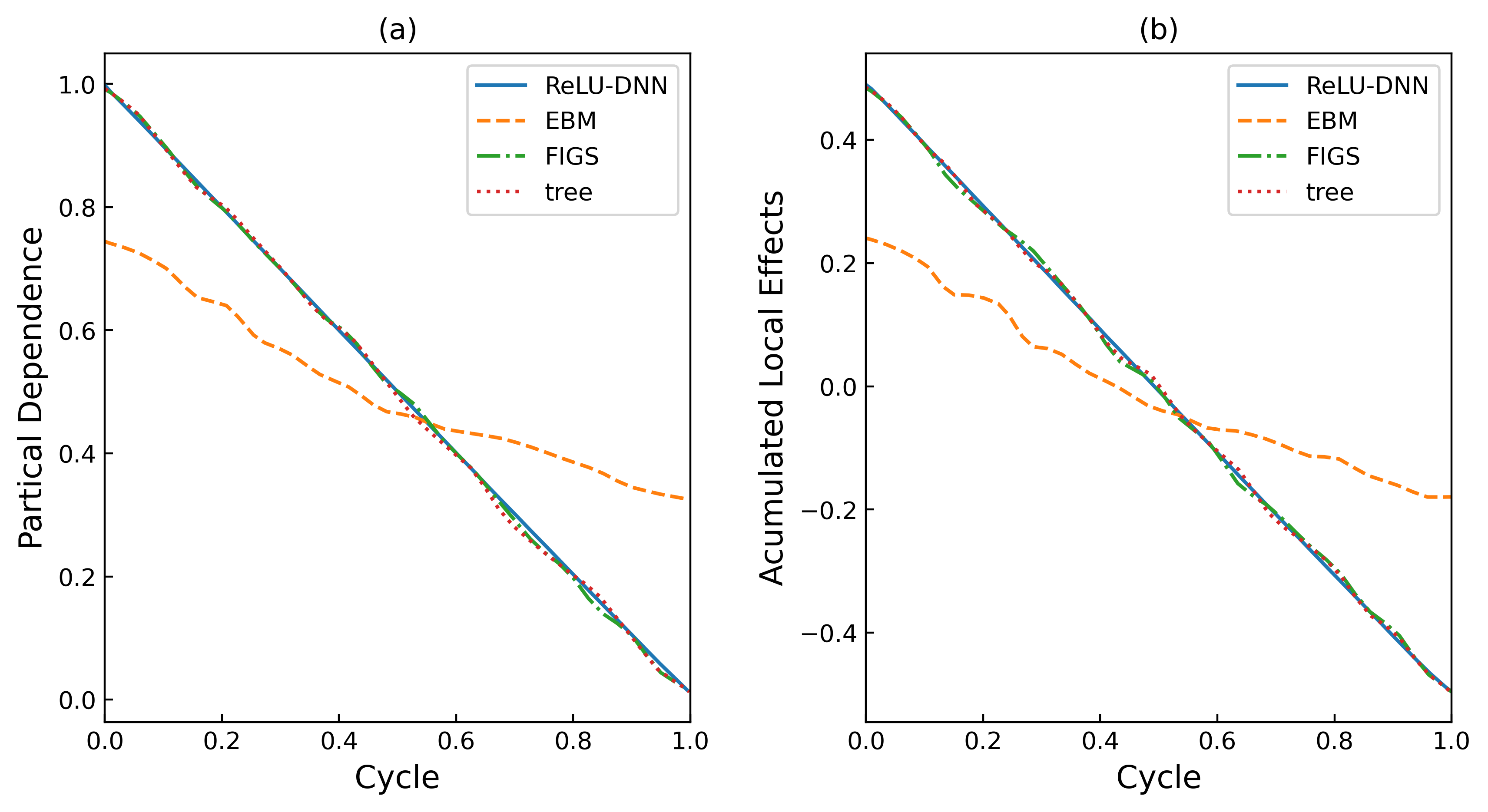}
\caption{(a) Univariate Partial dependence plot (PDP) and (b) accumulated local effects (ALE) for global model explainability: ReLU-DNN, EBM, FIGS, and tree.}
\label{fig:global}
\end{figure}

Figure~\ref{fig:global} (a) provides the global explainability of the model using the univariate partial dependence plot (PDP)  for global model explainability. PDPs are an interpretability ML method to understand the relationship between a single feature (or a small data set) and the target variable~\cite{law2007maturing}. PDPs can help determine whether a feature and its target variable have a linear or more complex relationship by demonstrating how a feature affects a model's predictions. They can also help identify potential confounding factors or traits that might obscure the connection between an intriguing quality and the desired result. PDPs can be useful when explaining a model's predictions to non-technical audiences because they are typically easy to understand. They are subject to some limitations, however. PDPs can only provide a partial picture of how the model behaves when a feature is linked with other features. Due to their sensitivity to the mean or median values used to set the values of the other features, PDPs may not accurately depict the model's behavior for extreme or out-of-range values of the feature.

\begin{comment}
\begin{figure}[H]
\begin{adjustwidth}{-\extralength}{0cm}
\centering
\subfloat[\centering][\centering]{\includegraphics[width=150mm]{reluGLOBALpdp.png}\label{reluGLOBALpdp}}\vspace{-6pt}\\
\subfloat[\centering][\centering]{\includegraphics[width=150mm]{ebmGLOBALpdp.png}\label{ebmGLOBALpdp}}\vspace{-6pt}\\
\subfloat[\centering][\centering]{\includegraphics[width=150mm]{figsGLOBALpdp.png}\label{figsGLOBALpdp}}\vspace{-6pt}\\
\subfloat[\centering][\centering]{\includegraphics[width=150mm]{treeGLOBALpdp.png}\label{treeGLOBALpdp}}\vspace{-6pt}\\
\end{adjustwidth}
\caption{Univariate Partial dependence plot (PDP) and accumulated local effects (ALE) for global model explainability:
(\textbf{a}) ReLU-DNN; 
(\textbf{b}) EBM;
(\textbf{c}) FIGS;
(\textbf{d}) tree.}\label{ModelExplainGlobalpdp}
\end{figure}
\end{comment}

\subsubsection{Global Explainability: Accumulated Local~Effects}
Figure~\ref{fig:global} (b) also uses accumulated local effects (ALE), a technique used in ML interpretability~\cite{law2007maturing}. The relationship between a single feature (or a small set of features) and the target variable in an ML model can be understood using ALE plots. An ALE plot is produced by plotting the model's average prediction as a function of the feature of interest while maintaining the observed values of all other features. ALE plots can show how a feature influences a model's predictions and reveal whether a feature and the target variable have a linear or more complex relationship. They can also aid in locating potential confounding variables or characteristics that obfuscate the link between an interesting feature and the desired outcome. The ability to depict more intricate relationships between the feature of interest and the target variable is one advantage ALE plots have over PDPs. Since they rely on the observed values rather than the mean or median used to fix the values of the other features, they are also less sensitive to this decision. ALE plots do, however, have some drawbacks. As they plot the model's prediction for each data point rather than the average prediction across many data points, they can be trickier to interpret than other techniques. Additionally, they might be sensitive to data noise and inaccurately depict the model's behavior for extreme or out-of-range values of the feature.

%LOCAL-LIME
\subsubsection{Local Explainability: Local Interpretable Model Agnostic~Explanations}
\begin{figure}[!htbp]
    \centering
    \includegraphics[width=\textwidth]{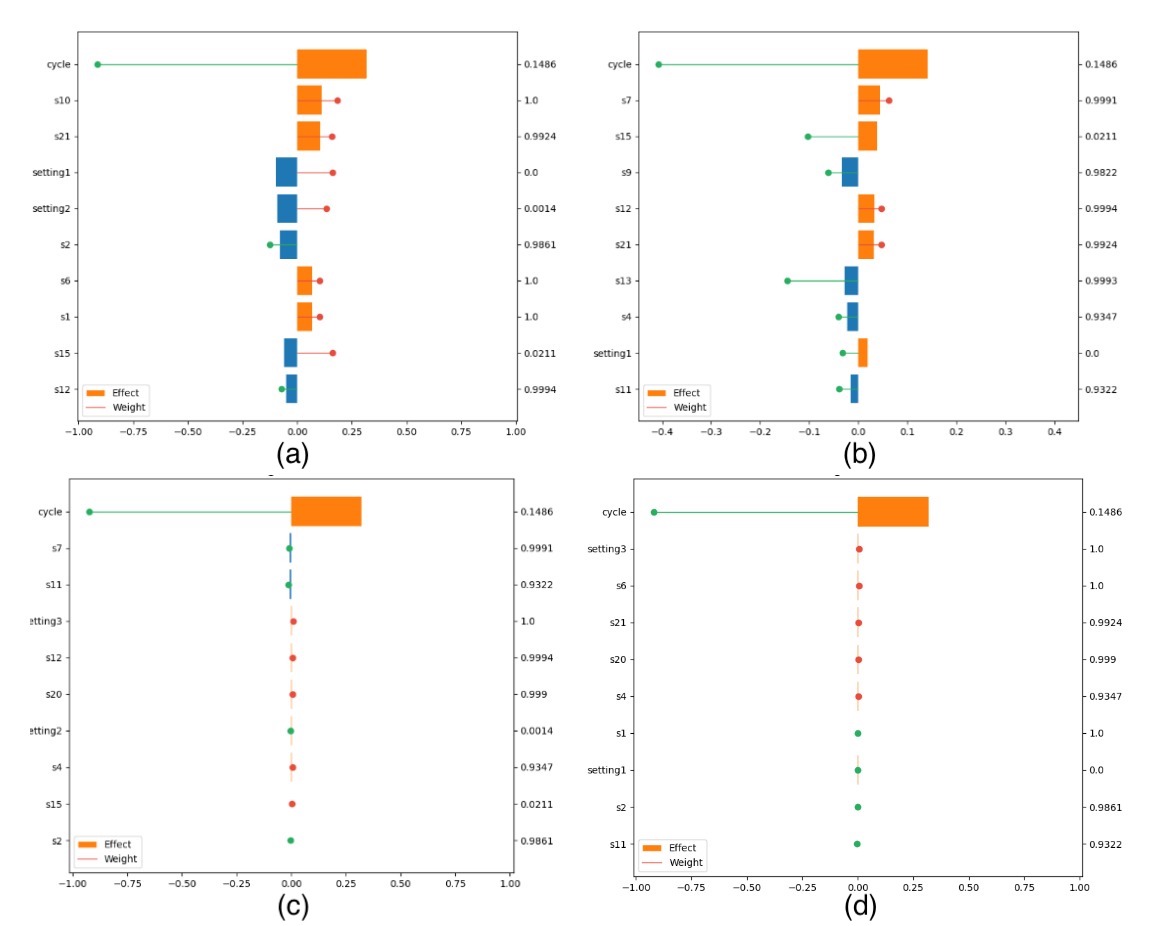}
    \caption{Local model explainability using LIME (a) ReLU-DNN, (b) EBM, (c) FIGS, and (d) Tree. }
\label{ModelExplainLocalLIME}
\end{figure}

Figure~\ref{ModelExplainLocalLIME} shows the local explainability and XAI outcome from local interpretable model-agnostic explanations (LIME)  \cite{ribeiro2016model}. LIME~\cite{benchekroun2020need}  explains complex, "black box" machine learning model predictions. It is model-agnostic and provides local explanations for a single data point or small set of data points. LIME can reveal a black box model's most important predictors and decision-making processes while identifying model biases and ensuring the model is not making predictions based on irrelevant or misleading features. However, it can potentially only explain a model's prediction for a small set of data points locally. It may not accurately represent model behavior. It may not always provide accurate or reliable explanations~\cite{khan2022explainable}. It is clear in Figure~\ref{ModelExplainLocalLIME} that the cycle governs the RUL in all four models. However,  ReLU-DNN and EBM have better expressibility than FIGS and tree algorithms, consistent with global explainability (Figure~\ref{ModelExplainGlobal}).

\begin{comment}
\begin{figure}[H]
\begin{adjustwidth}{-1 in}{0cm}
\centering
\subfloat[\centering][\centering]{\includegraphics[width=90mm]{reluLIME.png}\label{reluLIME}}
\subfloat[\centering][\centering]{\includegraphics[width=90mm]{ebmLIME.png}\label{ebmLIME}}\\
\end{adjustwidth}
%\caption{{\em Cont.}}
\label{ModelExplainLocalLIME}
\end{figure}

\begin{figure}[H]\ContinuedFloat
\centering
\begin{adjustwidth}{-1 in}{0cm}
\subfloat[\centering][\centering]{\includegraphics[width=90mm]{figLIME.png}\label{figLIME}}
\subfloat[\centering][\centering]{\includegraphics[width=90mm]{treeLIME.png}\label{treeLIME}}
\end{adjustwidth}
\caption{Local model explainability using LIME
(\textbf{a}) ReLU-DNN; 
(\textbf{b}) EBM;
(\textbf{c}) FIGS;
(\textbf{d}) tree.}
\label{ModelExplainLocalLIME}
\end{figure}
\end{comment}
\unskip

% %LOCAL-SHAP
\subsubsection{Local Explainability: SHapley Additive~Explanation}
To overcome the limitations of existing methods, Lundberg~et~al.~\cite{lundberg2017unified,lundberg2022unified} introduced the SHAP (shapley additive explanation)  technique. SHAP~\cite{kawakura2022analyses}  assigns credit or value to each feature for its contribution to machine learning model prediction. Shapley values from game theory underpin it. SHAP values are calculated by comparing a data point's model prediction to a reference set's prediction. Features unique to the data point of interest explain the difference between its prediction and the reference prediction. In a bar plot, the horizontal axis represents features, and the vertical axis represents SHAP values. SHAP values can help identify model biases and the most important predictors. They can also ensure the model predicts relevant and meaningful features. SHAP values are sensitive to the reference data set and may not provide accurate or reliable explanations.

\begin{figure}[!htbp]
\setcounter{subfigure}{0}
%\begin{adjustwidth}{-1in}{-1in} % Adjusts both left and right margins equally
\nointerlineskip\leavevmode
\centering
\subfloat[\centering]{\includegraphics[width=90mm]{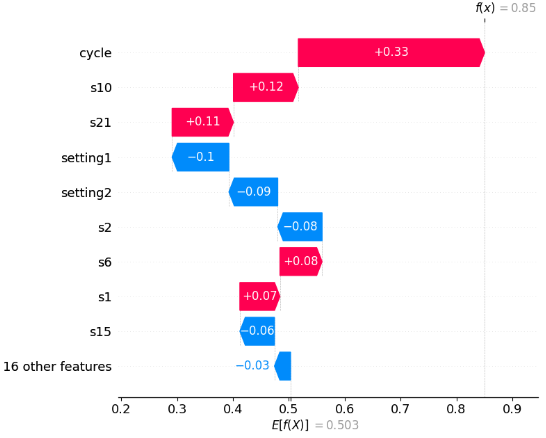}\label{reluSHAP}}
\subfloat[\centering]{\includegraphics[width=90mm]{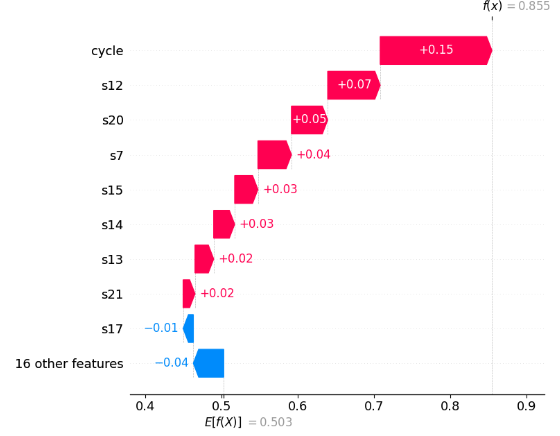}\label{ebmSHAP}}\\
\subfloat[\centering]{\includegraphics[width=90mm]{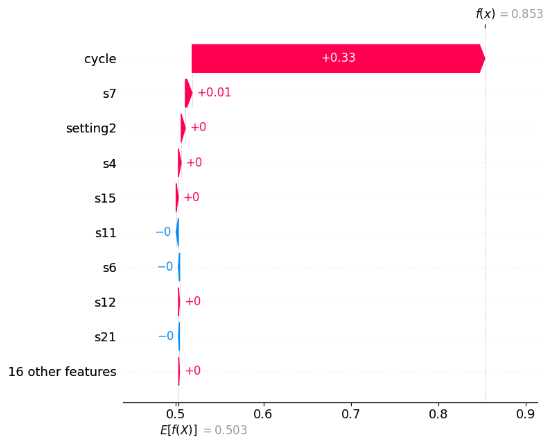}\label{figSHAP}}
\subfloat[\centering]{\includegraphics[width=90mm]{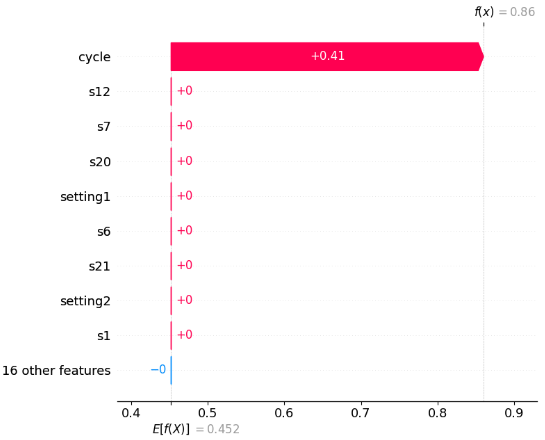}\label{treeSHAP}}
%\end{adjustwidth}
\caption{Local model explainability using SHAP
(\textbf{a}) ReLU-DNN; 
(\textbf{b}) EBM;
(\textbf{c}) FIGS;
(\textbf{d}) tree.}\label{ModelExplainLocalSHAP}
\end{figure}

In Figure~\ref{ModelExplainLocalSHAP}, SHAP values for a specific prediction made by a machine learning model are displayed using the sample with index zero. The~horizontal bar plot labeled "important scores" shows the Shapley value for each feature. The~impact of each feature on the model's final output is depicted graphically, with~the base value being the model's average output over the training dataset and the final value representing the prediction obtained from the sample of interest. According to this figure, it appears that the "cycle" feature has a significant impact on the model's prediction of RUL (remaining useful life). This is confirmed and explained by all of the models in the figure. Additionally, it can be seen that ReLU-DNN and EBM have better explanatory power than FIGS and tree algorithms. This is consistent with the findings in Figures~\ref{ModelExplainGlobal} and~\ref{ModelExplainLocalLIME}.

\subsection{Interpretability of the~Model}
%%%%%%%%%%%%%%%%%%%%%%%%%%%%%%%%%%%%%%%%%%%%%%%%%%%%%%%%%%%%%
% https://arxiv.org/pdf/2111.01743.pdf
\subsubsection{Local~Interpretability}

\begin{figure}[!htbp]
\centering % Ensure global centering
\subfloat[\centering]{\includegraphics[width=70mm, height=40mm]{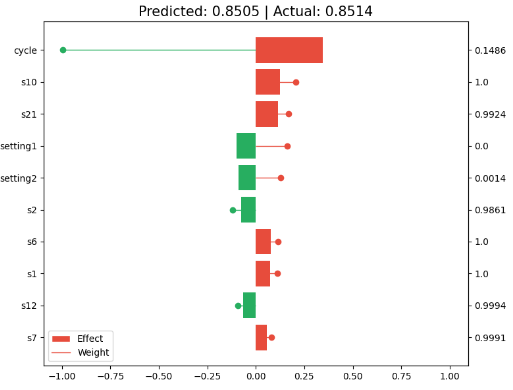}\label{reluLOCALinterpret}}\\
\vspace{-12pt} % Adjust vertical space as needed
\subfloat[\centering]{\includegraphics[width=125mm, height=40mm]{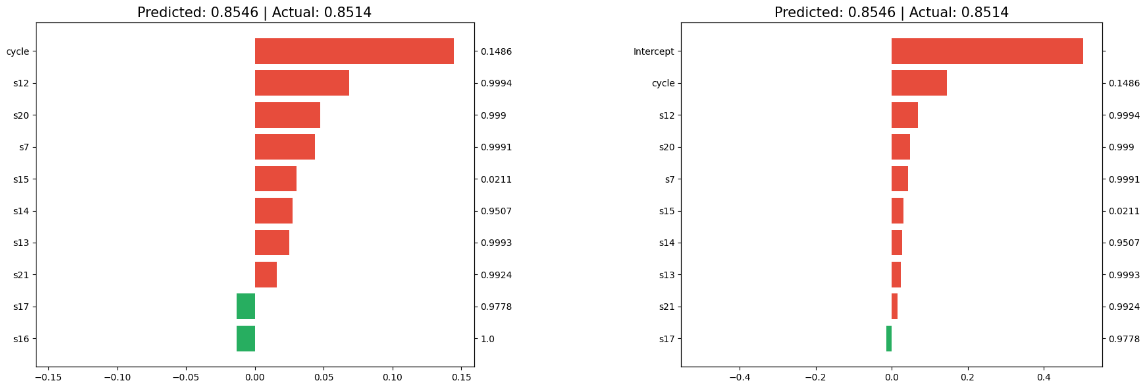}\label{ebmLOCALinterpret}}\\
\vspace{-12pt} % Adjust vertical space as needed
\subfloat[\centering]{\includegraphics[width=85mm, height=45mm]{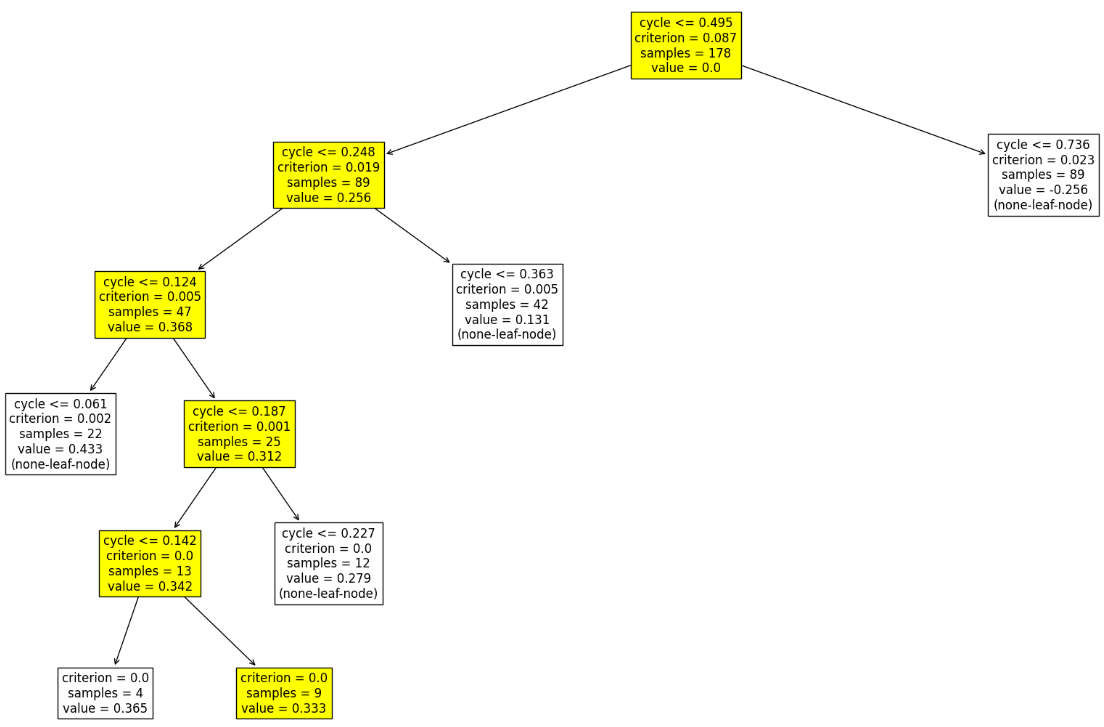}\label{figLOCALinterpret}}\\
\vspace{-12pt} % Adjust vertical space as needed
\subfloat[\centering]{\includegraphics[width=85mm, height=45mm]{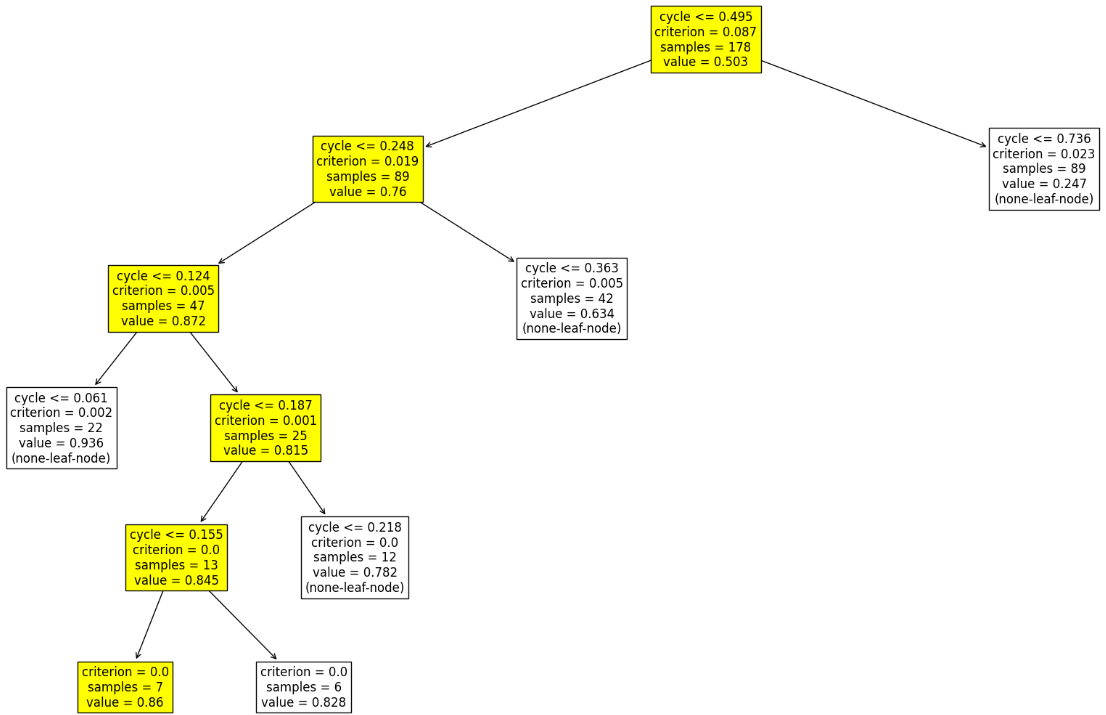}\label{treeLOCALinterpret}}
\caption{Local Model interpretability:
(\textbf{a}) ReLU-DNN; 
(\textbf{b}) EBM;
(\textbf{c}) FIGS;
(\textbf{d}) Tree.}\label{ModelInterpretLocal}
\end{figure}

 Figure~\ref{ModelInterpretLocal} shows the local exact interpretability in terms of weight and effect. Local interpretability generally refers to the ability to explain the ML predictions for a specific data point rather than the whole model. The~weight and effect of a feature can provide insight into the most important prediction factors. However, the~weight and effect of a feature are specific to the individual data point being explained and may not be generalizable to other data points or the model as a whole. Figure~\ref{ModelInterpretLocal}a shows the local exact interpretability in terms of weight and effect of ReLU-DNN. It shows that the cycle significantly impacts the RUL prediction. Figure~\ref{ModelInterpretLocal}b shows the local interpretability of EBM for local feature importance (left in Figure~\ref{ModelInterpretLocal}b) and local effect importance (right in Figure~\ref{ModelInterpretLocal}b). Both the local feature and effect confirm the impact of the cycle on RUL prediction. Figure~\ref{ModelInterpretLocal}c,d show the local interpretability of FIGS and tree. Both algorithms confirm the relationship between the cycle and~RUL.

% Figure 13

\unskip

\subsubsection{Global~Interpretability}
\begin{figure}[!htbp]
    \centering
    \includegraphics[width=\textwidth, height=12cm]{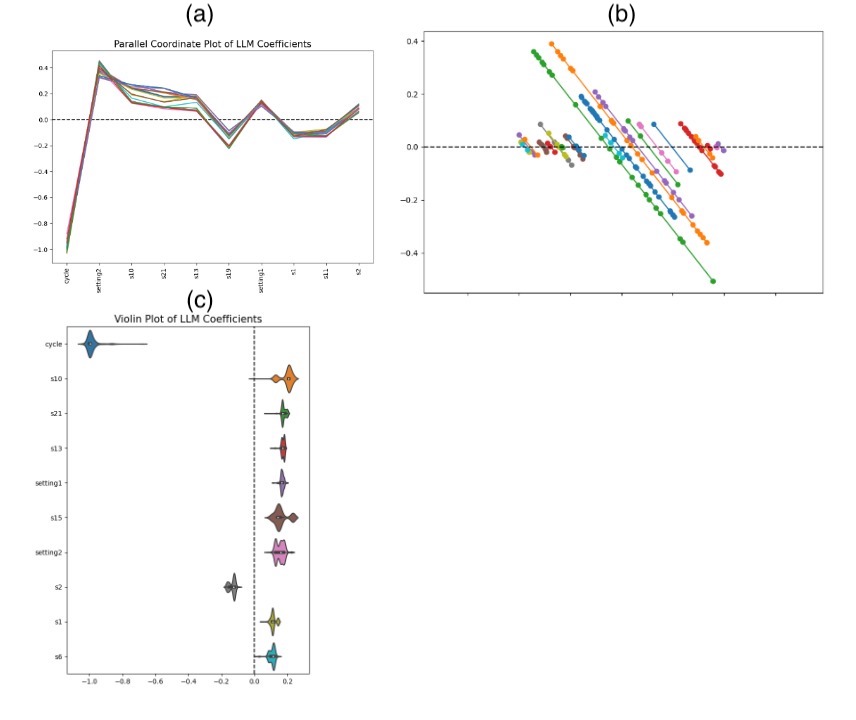}
    \caption{(a) Parallel coordinate plot of the LLM coefficient, (b) LLM feature importance of the cycle, and (c) Violin plot of LLM coefficient.}
    \label{ReLUModelInterpretGLOBAL}
\end{figure}

Figure~\ref{ReLUModelInterpretGLOBAL} shows the global interpretability of ReLU-DNN only, and~the rest of the other model's interpretability is provided in Appendix \ref{app1}  (Figure~\ref{ModelInterpretGLOBAL}). Global Interpretability has been performed following the approach by~\cite{sudjianto2020unwrapping} and explained in terms of the parallel coordinate plots of local linear models (LLMs) coefficients and LLM feature~importance.

Figure~\ref{ReLUModelInterpretGLOBAL}a shows the parallel coordinate plot of the  LLMs coefficient for ReLU-DNNs, which can be used to visualize the contribution of each feature to the model's prediction, and the relative magnitudes and directions of its effects. For~example, consider a ReLU-DNN with three input features: $x_{1}$, $x_{2}$, and~$x_{3}$. If~the parallel coordinate plot for this ReLU-DNN displays the coefficients for each feature on separate vertical axes, then the line connecting the axes would represent the weight and direction of the effect of each feature on the model's prediction. If~the line is shallow and slopes downwards, it indicates a weak negative effect. The~position of the line along the axis also reflects the magnitude of the effect: a line that is higher on the axis indicates a larger effect, whereas a line that is lower on the axis indicates a smaller effect. Parallel coordinate plots of LLM coefficients can be useful for identifying patterns or trends in the data and comparing the relative importance of different features. However, they can be difficult to interpret when the number of features is large, as~the lines can overlap and the plot can become cluttered. It is important to remember that these plots only represent the local linear behavior of the ReLU-DNN and may not accurately reflect the overall behavior of the model. Figure~\ref{ReLUModelInterpretGLOBAL}b shows the LLM feature importance. It refers to each feature's influence on the RUL prediction.  These two interpretability methods conclude that cycle-representing profile plots of the most important feature for RUL prediction, and~it has been confirmed by the violin plot of LLM coefficient for ReLU-DNN, as~seen in Figure~\ref{ReLUModelInterpretGLOBAL}c. 

\begin{comment}
The LLM summary of ReLU-DNN is shown in Appendix \ref{app1} as Table~\ref{LLMsummary}.
\end{comment}

\begin{comment}
\begin{figure}[H]
\subfloat[\centering][\centering]{\includegraphics[width=85mm]{reluGLOBALLM.png}\label{reluGLOBALLM}}\\
\caption{\textit{Cont}.}\label{ReLUModelInterpretGLOBAL}
\end{figure}
\begin{figure}[H]\ContinuedFloat
\subfloat[\centering][\centering]{\includegraphics[width=85mm]{reluGLOBALLMFeture.png}\label{reluGLOBALLMFeture}}\\
\subfloat[\centering][\centering]{\includegraphics[width=65mm]{reluInterpretViolin.png}\label{reluInterpretViolin}}%MDPI: Please explain all colors.
\caption{\hl{Global} Model interpretability of ReLU-DNN:
(\textbf{a}) Parallel coordinate plot of the LLM coefficient.
(\textbf{b}) LLM feature importance of the cycle.
(\textbf{c}) Violin plot of LLM coefficient.}\label{ReLUModelInterpretGLOBAL}
\end{figure}
\end{comment}

\unskip

\subsection{Trustworthy AI: Model Diagnosis and Validation}
\begin{figure}[htbp]
\setcounter{subfigure}{0}
\centering
\subfloat[\centering][\centering]{\includegraphics[scale=0.4]{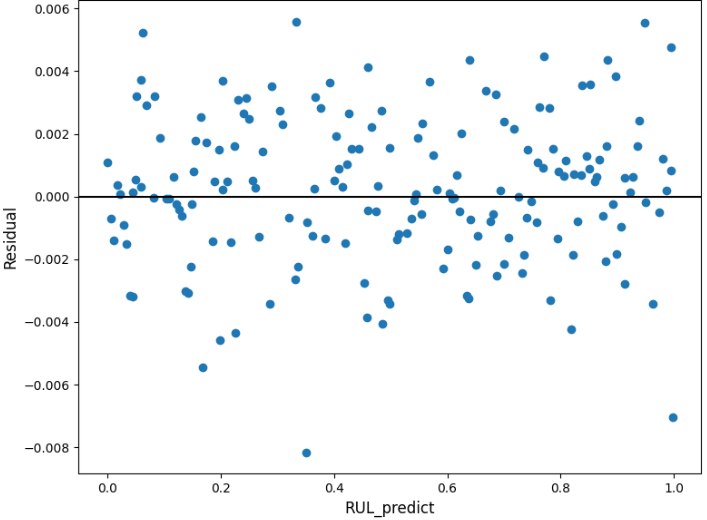}\label{reluTRAINdiagnose}}
\subfloat[\centering][\centering]{\includegraphics[scale=0.4]{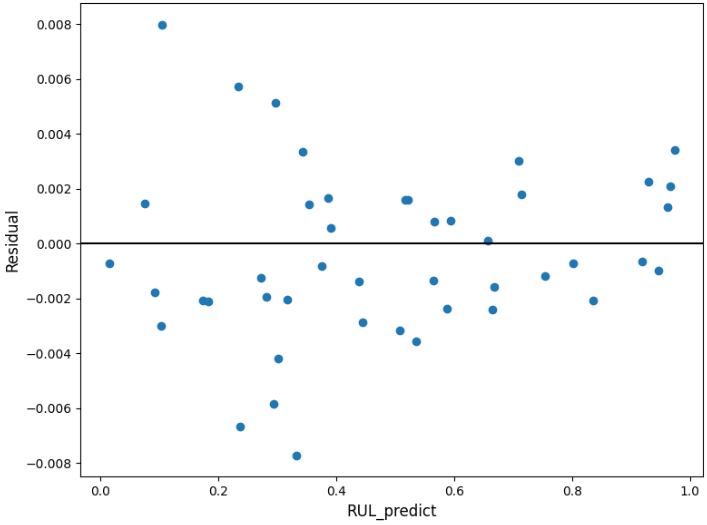}\label{reluTESTdiagnose}}\vspace{-6pt}\\
\subfloat[\centering][\centering]{\includegraphics[scale=0.4]{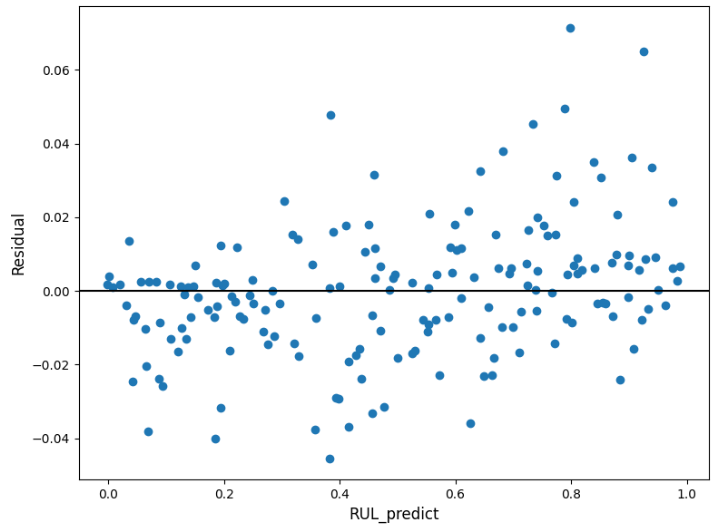}\label{ebmTRAINdiagnose}}
\subfloat[\centering][\centering]{\includegraphics[scale=0.4]{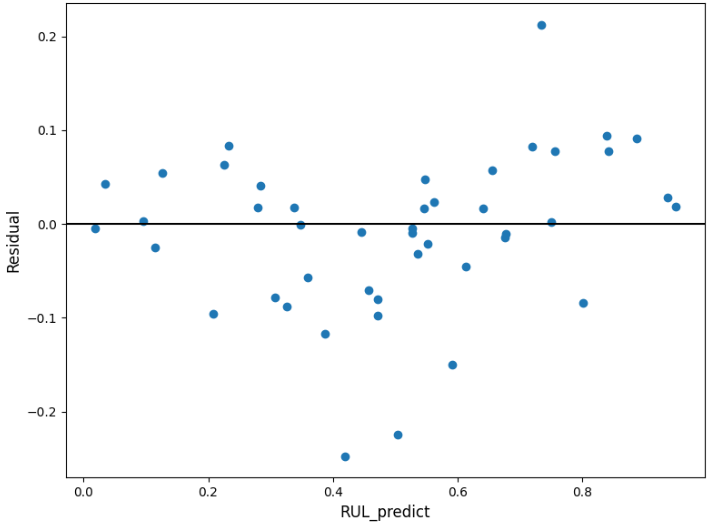}\label{ebmTESTdiagnose}}\vspace{-6pt}\\
\subfloat[\centering][\centering]{\includegraphics[scale=0.4]{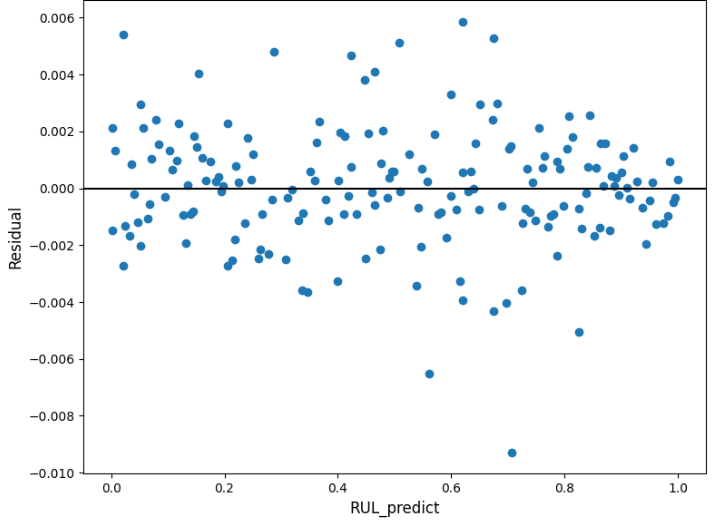}\label{figsTRAINdiagnose}}
\subfloat[\centering][\centering]{\includegraphics[scale=0.4]{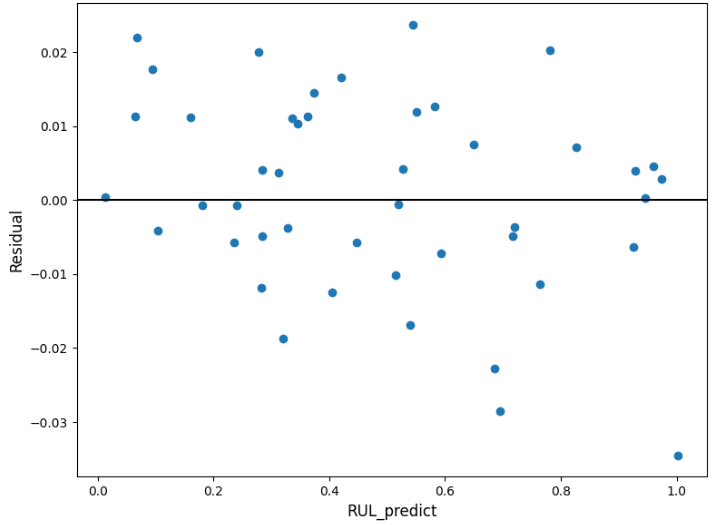}\label{figsTESTdiagnose}}\vspace{-6pt}\\
\subfloat[\centering][\centering]{\includegraphics[scale=0.4]{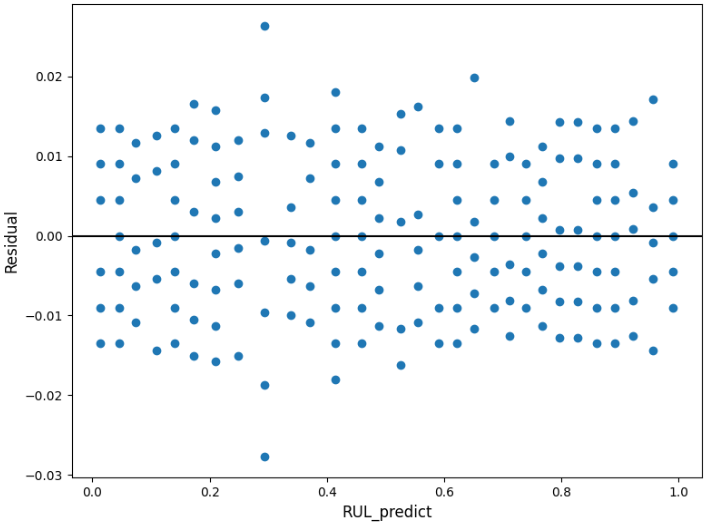}\label{treeTRAINdiagnose}}
\subfloat[\centering][\centering]{\includegraphics[scale=0.4]{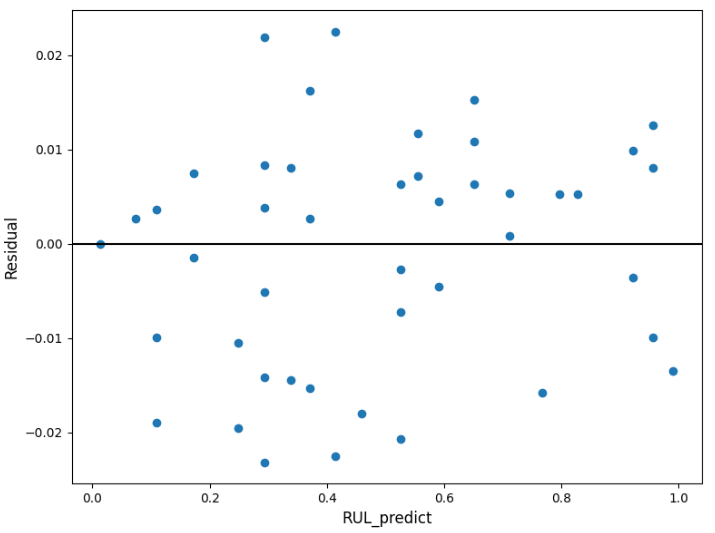}\label{treeTESTdiagnose}}\vspace{-6pt}
\caption{Residual Plot for RUL prediction:
(\textbf{a}) ReLU-DNN training dataset; 
(\textbf{b}) ReLU-DNN testing dataset; 
(\textbf{c}) EBM training dataset;
(\textbf{d}) EBM testing dataset;
(\textbf{e}) FIGS training dataset;
(\textbf{f}) FIGS testing dataset;
(\textbf{g}) tree training dataset;
(\textbf{h}) tree testing dataset. }\label{ModelDiagnoseRUL}
\end{figure}

%%%%%%%%%%%%%%%%%%%%%%%%%%%%%%%%%%%%%%%%%%%%%%%%%%%%%%%%%%%%%
The diagnosis outcomes of candidate models are presented in terms of accuracy, overfit, reliability, robustness, and resilience. Table~\ref{Accuracy} exhibits the model accuracy in terms of MSE, MAE, and~R2 performance evaluation. Additionally, Figure~\ref{ModelDiagnoseRUL} exhibits the residual plot for RUL prediction for training and testing data, which clarifies the model's performance, as~shown in Table~\ref{Accuracy}. ReLU-DNN exhibited the best performance in terms of all the evaluation matrices, followed by tree and FIGS, whereas EBM exhibits inferior accuracy compared to other~candidates.

%%%%%%%%%
\begin{table}[!htbp]
\centering
\caption{Model Accuracy}
\label{Accuracy}
\begin{adjustbox}{width=\textwidth}
\begin{tabular}{@{}llllllll@{}}
\toprule
\multirow{2}{*}{Accuracy} &
  \multicolumn{3}{c}{Tree} &
  \multirow{2}{*}{Accuracy} &
  \multicolumn{3}{c}{EBM} \\ \cmidrule(lr){2-4} \cmidrule(l){6-8} 
 &
  \multicolumn{1}{c}{MSE} &
  \multicolumn{1}{c}{MAE} &
  \multicolumn{1}{c}{R2} &
   &
  \multicolumn{1}{c}{MSE} &
  \multicolumn{1}{c}{MAE} &
  \multicolumn{1}{c}{R2} \\ \midrule
Train &
  0.00001 &
  0.0083 &
  0.9989 &
  Train &
  0.0003 &
  0.0133 &
  0.9961 \\
Test &
  0.00001 &
  0.0102 &
  0.9980 &
  Test &
  0.0070 &
  0.0607 &
  0.9054 \\
Gap &
  1.00000 &
  0.0018 &
  -0.0009 &
  Gap &
  0.0066 &
  0.0474 &
  -0.0907 \\ \midrule
\multirow{2}{*}{Accuracy} &
  \multicolumn{3}{c}{FIGS} &
  \multirow{2}{*}{Accuracy} &
  \multicolumn{3}{c}{ReLU-DNN} \\ \cmidrule(lr){2-4} \cmidrule(l){6-8} 
 &
  \multicolumn{1}{c}{MSE} &
  \multicolumn{1}{c}{MAE} &
  \multicolumn{1}{c}{R2} &
   &
  \multicolumn{1}{c}{MSE} &
  \multicolumn{1}{c}{MAE} &
  \multicolumn{1}{c}{R2} \\ \midrule
Train &
  0.00000 &
  0.0015 &
  0.9999 &
  Train &
  0.00000 &
  0.0018 &
  0.9999 \\
Test &
  0.00002 &
  0.0104 &
  0.9977 &
  Test &
  0.00000 &
  0.0025 &
  0.9999 \\
Gap &
  0.00002 &
  0.0089 &
  -0.0023 &
  Gap &
  0.00000 &
  0.0006 &
  -0.0001 \\ \bottomrule
\end{tabular}
\end{adjustbox}
\end{table}

\begin{figure}[H]
\centering
\includegraphics[width=\textwidth]{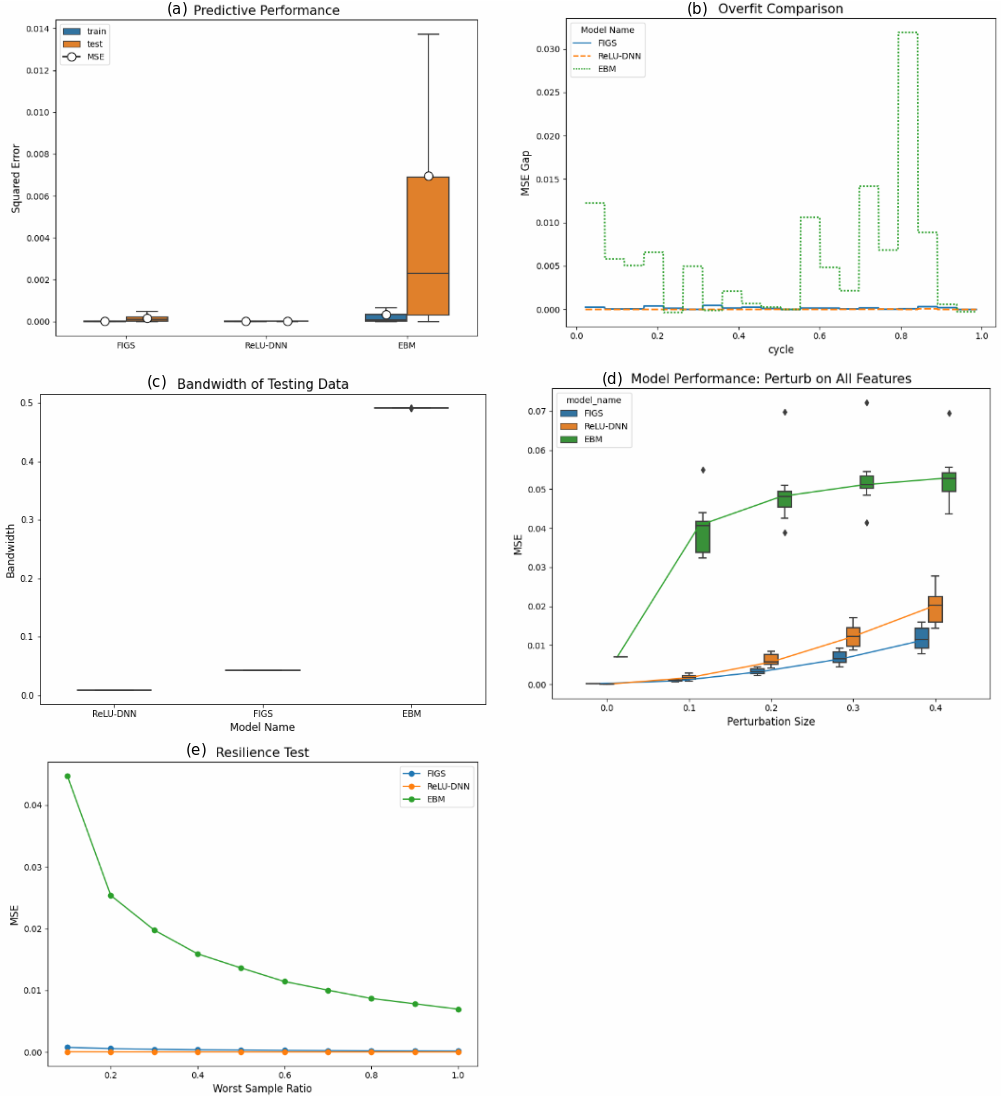}
\caption{Model prediction performance comparison: (a) accuracy, (b) overfit, (c) reliability, (d) robustness, and (e) resilience.}
\label{CompareModelPerformance}
\end{figure}

 Accuracy is provided in terms of MSE, MAE, and~R2 in Figure~\ref{CompareModelPerformance}a. It is clear that  ReLU-DNN and FIGS significantly outperformed the EBM algorithm. It is also exhibited in Figure~\ref{CompareModelPerformance}a that ReLU-DNN exhibited superior predictive performance. Figure~\ref{CompareModelPerformance}b shows the Overfitting, which refers to differences between "true" and "fitting" errors, which can be used to evaluate overfitting: the larger the gap, the more overfitting there is. EBM shows a higher overfitting tendency, mostly due to the large MSE gap between the training and testing dataset, as seen in Table~\ref{Accuracy}. Figure~\ref{CompareModelPerformance}c shows the reliability test using split conformal prediction, segmented bandwidth, and~distribution shift (reliable vs. unreliable), among which robustness can explain in Figure~\ref{CompareModelPerformance}d. A robustness test uses covariate perturbation and worst sampler resilience. It can be seen in Figure~\ref{CompareModelPerformance}d  that in the event of perturbation occurring in all features, EBM exhibits inferior predictive performance to the other candidate algorithms. The~outcomes are also consistent with the "resilience test", which exhibits prediction performance degradation using worst-case subsampling and under out-of-distribution scenarios, as~seen in Figure~\ref{CompareModelPerformance}e.

\begin{comment}
\begin{figure}[H]
\begin{adjustwidth}{-\extralength}{0cm}
\centering
\subfloat[\centering][\centering]{\includegraphics[width=85mm]{PredictPerformAccuracy.png}\label{PredictPerformAccuracy}}
\subfloat[\centering][\centering]{\includegraphics[width=85mm]{PredictPerformOverfit.png}\label{PredictPerformOverfit}}
\end{adjustwidth}
\caption{\textit{Cont}.}\label{CompareModelPerformance}
\end{figure}

\begin{figure}[H]\ContinuedFloat
\begin{adjustwidth}{-\extralength}{0cm}
\centering
\subfloat[\centering][\centering]{\includegraphics[width=85mm]{PredictPerformReliability.png}\label{PredictPerformReliability}}
\subfloat[\centering][\centering]{\includegraphics[width=85mm]{PredictPerformRobustness.png}\label{PredictPerformRobustness}}\\
\subfloat[\centering][\centering]{\includegraphics[width=85mm]{PredictPerformResilience.png}\label{PredictPerformResilience}}
\end{adjustwidth}
\caption{Model prediction performance comparison:
(\textbf{a}) accuracy; 
(\textbf{b}) overfit; 
(\textbf{c}) reliability; 
(\textbf{d})~robustness;
(\textbf{e}) resilience.}\label{CompareModelPerformance}
\end{figure}
\end{comment}

%%%%%%%%%%%%%%%%%%%%%%%%%%%%%%%%%%%%%%%%%%
\section{Conclusions}
This study demonstrates that explainable and interpretable AI is indispensable for engendering reliable remaining useful life (RUL) predictions within intelligent digital twin frameworks. By elucidating the rationale and underscoring model decisions, explainability and interpretability mechanisms foster trust, transparency, and accountability in AI systems. The research presented herein advances the state-of-the-art by developing rigorous methodologies integrating cutting-edge explainable AI techniques to enhance the fidelity of RUL forecasts generated by digital twins across safety-critical applications. 

This paper justified the importance of XAI and INL for digital twin components and leveraged state-of-the-art XAI and interpretable ML (IML) algorithms, ReLU-DNN, EBM, FIGS, and ~decision tree, to ensure the use of trustworthy AI/ML applications for useful life (RUL) prediction. The~explainability and interpretability were observed for both local and global aspects while demonstrating the fact that the longevity of the cycle operation mostly impacts RUL prediction. It is also seen that the operating cycle feature is consistently shown to be the most influential factor governing remaining useful life across the XAI analysis, aligning with domain knowledge. Both model-agnostic and intrinsic explainability techniques conclusively demonstrate the predominance of the operating cycle feature in governing asset longevity across diverse model architectures. These findings conceptually validate the capability of explainable AI to elucidate the causal mechanisms driving equipment degradation, unlocking invaluable insights into optimal maintenance and operational procedures. By combining trustworthy and interpretable ML with physics-based digital twins, this research enables a path toward responsible and ethical AI automation that augments human decision-making for industrial prognostic health management.

Future work will focus on specific aspects of realistic digital twin for monitoring and control:

\begin{itemize}
\item Develop an explainable and accelerated prediction algorithm that can handle multi-fidelity and scarce data, missing or approximate physics, and generalize to new environments. This will enable real-time inference for the digital twin.

\item Create algorithms to model sensor location optimization, sensor degradation, re-calibration, and sensor signal reconstruction ~\cite{kabir2010non} to understand their impact on overall system degradation predictions.

\item Build an online learning algorithm that can continuously update the digital twin models and parameters to maintain temporal synchronization with the physical asset.

\item Create uncertainty quantification methods ~\cite{kumar2020uncertainty} tailored for scarce, noisy datasets and never-before-seen noisy data. The methods should account for spatiotemporal degradation patterns.

\end{itemize}

\appendix
\section{Global Model Interpretability}
\label{app1}

\begin{figure}[H]
\setcounter{subfigure}{0}
\centering
\subfloat[\centering]{\includegraphics[width=75mm]{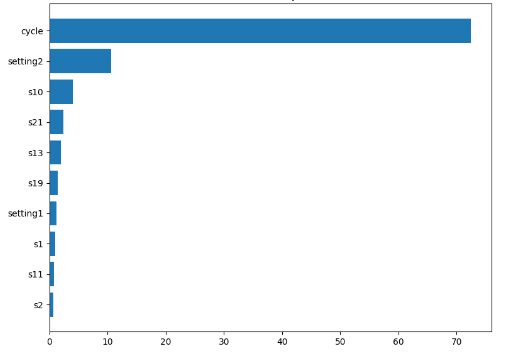}\label{reluDNNFeatureimportance}}\\
\subfloat[\centering]{\includegraphics[width=120mm]{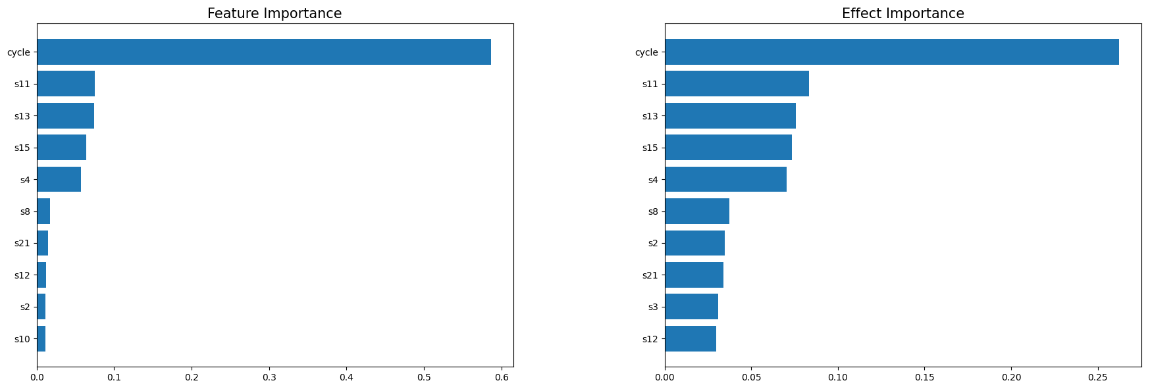}\label{ebmGLOBALinterpret}}\\
\subfloat[\centering]{\includegraphics[width=100mm]{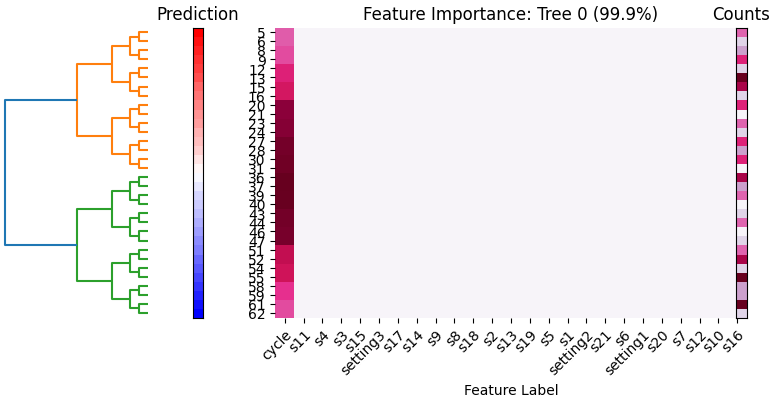}\label{figGLOBALinterpret}}
\caption{Global Model interpretability:
(\textbf{a}) LLM Feature importance of ReLU-DNN; 
(\textbf{b}) Feature and effect importance of EBM;
(\textbf{c}) Feature importance of FIGS.}\label{ModelInterpretGLOBAL}
\end{figure}

\section*{Acknowledgement}
The computational part of this work was supported in part by the National Science Foundation (NSF) under Grant No. OAC-1919789.

\section*{Declaration of Generative AI and AI-assisted technologies in the writing process}
During the preparation of this work the author(s) used GPT-4 (Generative Pre-trained Transformer 4)  in order to language editing and refinement. After using this tool/service, the author(s) reviewed and edited the content as needed and take(s) full responsibility for the content of the publication [\href{https://www.elsevier.com/about/policies/publishing-ethics/the-use-of-ai-and-ai-assisted-writing-technologies-in-scientific-writing}{Elsevier Publishing Ethics}].


\bibliographystyle{unsrtnat}
\bibliography{references}  %%% Uncomment this line and comment out the ``thebibliography'' section below to use the external .bib file (using bibtex).





% \end{thebibliography}
\clearpage

\end{document}